\documentclass[review,12pt]{elsarticle}
\journal{Future Generation Computer Systems}

\usepackage{amsmath}
\usepackage{amsfonts}
\usepackage{amssymb}
\usepackage{algorithm}
\usepackage{algorithmic}
\usepackage{graphicx}
\usepackage{booktabs}
\usepackage{array}
\usepackage{url}
\usepackage{hyperref}
\graphicspath{{figures/}}

\begin{document}

\begin{frontmatter}

\title{Four conditions for composing plugin-based agents without a host}

\author[1]{Hanzhang Jia\corref{cor1}}
\ead{jh303@sussex.ac.uk}
\author[1]{Liheng Zeng}
\author[2]{Hao Cheng}
\author[3]{Yi Gao}
\author[2]{Bo Ma}
\author[2]{Guolong Xu}
\cortext[cor1]{Corresponding author}

\affiliation[1]{organization={University of Sussex}, city={Brighton}, postcode={BN1 9RH}, country={United Kingdom}}
\affiliation[2]{organization={Zhejiang Gongshang University}, city={Hangzhou}, postcode={310018}, country={China}}
\affiliation[3]{organization={Shanghai Shuyuan Information Technology Co., Ltd.}, city={Shanghai}, postcode={200000}, country={China}}

\begin{abstract}
Large language model agents now run long-horizon, multi-tool, autonomous sessions, and the infrastructure must keep the current assembly of capabilities, the progress of execution, and the basis for recovery after a fault. A single-process host holds this information in its own running state, so its death takes the session record and the assembly with it. This paper gives four conditions under which a plugin-based agent needs no host process, one each on state placement, record carriers, recovery coordination, and provider resolution. The store, writer, and delivery disciplines keep the conditions in force, and each carries formal support from the spatiotemporal-composability calculus and from the proofs in the supplementary materials. An agent that meets them can be constructed with its assembly on a bus of peer processes and its record in an append-only carrier owned by no process, and removing any one part makes the assembly need a host again. The conditions themselves are the contribution, and every implementation that meets them inherits the same advantages. Such an agent keeps its session through the death of any process, wastes no calls after a fault, loses no records, recovers locally, changes its capability set while it runs, and carries its undo semantics across processes. This paper builds one SDK of the conditions, Logos, and measures whether it fulfills their promise against the single-process reference and a configuration over MCP under one fault in each of 200 tasks, realized in 197. The reference lost all 197 faulted sessions, the MCP configuration kept its sessions but issued 1,212 wasted calls, and the SDK kept every session, wasted no call, lost no record, and fulfilled the promise of the conditions in every faulted task, with task performance preserved. The conditions give builders a tested route from single-process hosts to peer processes.
\end{abstract}


\end{frontmatter}

\section{Introduction}
\label{sec:intro}

Large language model agents are moving from single-question answering to long-horizon, multi-tool, multi-process, and autonomous operation~\cite{dsh2026spatiotemporal}. Coding assistants load tools to work on a repository~\cite{jimenez2024swebench,yang2024sweagent}. Orchestration frameworks route a task through the agents and tools assembled for that job~\cite{qu2025toolsurvey,yao2023react,wu2024autogen,hong2024metagpt}. Autonomous systems run many tools together through one long session~\cite{li2026agentharness}, and self-evolving agents create and refine their own tools as they operate~\cite{gao2026selfevo}. The runtime beneath the model prepares the input, forwards the calls, keeps the results, and handles recovery~\cite{li2026agentharness}.

A long session of this kind requires the infrastructure to keep, at all times, the current assembly of capabilities, the progress of execution, and the basis for recovery after a fault. Two requirements recur across these settings~\cite{cemri2025mast}. The session must survive the faults of its infrastructure, and the capability set must grow and shrink while the system runs, with every mounting and unmounting leaving a record that outlives the process that performed it. A single-process host holds this information in its own running state, so its death takes the session record and the assembly with it, and everything it carries shares one failure domain.

Existing agent infrastructure keeps this information inside a host. The Model Context Protocol (MCP) moves tool servers into processes of their own, while the composition of the agent and its session state stay in the host~\cite{hou2026mcp}, and agent-to-agent protocols delegate tasks among agents whose internal assembly stays private~\cite{ehtesham2025}. Durable-execution engines and hosted agent services persist an append-only event history and resume an interrupted run from it~\cite{temporal2025docs,anthropic2026managed}, and dynamic software updating replaces code without a restart yet leaves the assembly and the session records in the host~\cite{hicks2005dsu}. In each of these systems the record and its recovery stay inside one process, one library, or one platform, and none of them carries the assembly itself, together with the means to undo it, out of the host.

One of the formal supports of this paper is the spatiotemporal-composability calculus~\cite{dsh2026spatiotemporal}. It treats composition in two directions, a component mounted together with a recorded means of undoing its effect, so that unmounting reverts every side effect, and a component declaring its dependencies and finding their providers. The calculus proves the assembly recoverable when the components and the records stay in the one process that runs the agent, and once they fall into different processes it gives no formal guarantee. The question this paper answers is therefore whether the assembly and its records can leave the host process while keeping this semantics.

\begin{figure}
\centering
\includegraphics[width=\linewidth]{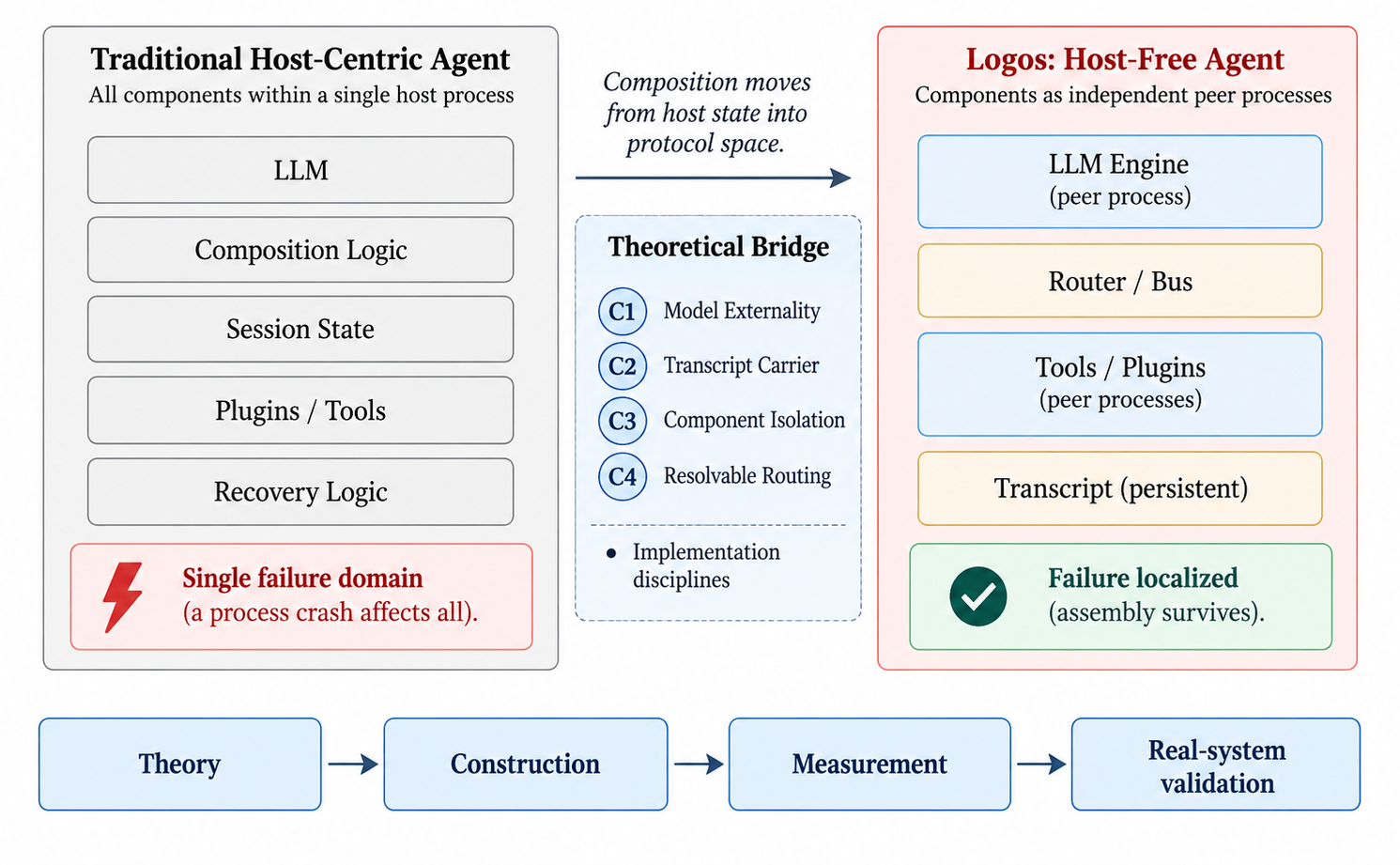}
\caption{The contrast between host-owned and host-free composition. The conditions of Section~\ref{sec:model} construct the host-free form. The route of the paper runs from the conditions through the SDK to measurement on a real system.}
\label{fig:overview}
\end{figure}

This paper answers with the conditions. Four engineering conditions, on state placement, record carriers, recovery coordination, and provider resolution, supported by the store, writer, and delivery disciplines, construct a plugin-based agent with no host process at all, and each condition carries formal support, the proofs in the supplementary materials and the semantics of the calculus among them. The construction reads in both directions, since removing any one part of it makes the assembly need a host again (the supplementary materials, S.3). The contribution of this paper is the conditions themselves, Logos is one SDK built on them, and the measurements prove this SDK fulfilling their promise. This paper builds one SDK of the conditions, Logos, which places the capability assembly on a bus shared by peer processes, following the peer-process paradigm of the Robot Operating System (ROS)~\cite{quigley2009ros}, and keeps the session record in an append-only carrier that no process exclusively owns. When a session process dies, a replacement process rebuilds the session from the record and continues it, so a process death removes only the execution running in that process. A plugin is a process written in any language that speaks the protocol, and the bus itself is an ordinary process that can be killed and replaced, so the assembly runs with no host at all. An agent built this way carries five advantages, it keeps its session through the death of any process, it wastes no calls and loses no records after a fault, it recovers locally to the faulted process, it changes its capability set while the system runs, and it carries its undo semantics across processes. Figure~\ref{fig:overview} contrasts the single-process host with this host-free form.

The evaluation compares Logos with two baselines under the same tasks, model, tool set, and fault action, the single-process reference implementation of the calculus~\cite{dsh2026repo} and a cross-process configuration over MCP. One fault was scheduled in each of 200 benchmark tasks and realized in 197. The single-process reference lost all 197 faulted sessions, the MCP configuration kept its sessions but issued 1,212 wasted calls on a dead endpoint, and the SDK kept every session, with zero wasted calls, zero lost records, and task performance preserved.

This paper makes three contributions.
\begin{itemize}
\item The conditions of host-free composition. Four conditions with their supporting disciplines make a plugin-based agent constructible with no host process, each condition carrying formal support, and the construction reads in both directions, so removing any one part makes the assembly need a host again (Section~\ref{sec:model}, supplementary materials S.1 through S.3). The conditions themselves are the contribution, and any implementation that meets them carries the same advantages.
\item One SDK of the conditions. Logos realizes the conditions construct by construct on a peer-process bus, the paradigm of the Robot Operating System. A plugin is a process, the router holds only a routing table rebuildable from broadcasts, and the session state needed for recovery lives in an append-only transcript that no process owns (Section~\ref{sec:construction}).
\item The evidence of fulfillment. The evaluation audits the conditions on every measured path and measures the SDK against both baselines under injected faults, on long-horizon delegated missions, and under planned changes of the capability set (Sections~\ref{sec:framework} and~\ref{sec:results}). Zero wasted calls and zero lost records hold for the SDK in every faulted task under both models, and neither baseline shares them. The failure domain narrows to the process that carries the fault.
\end{itemize}

Section~\ref{sec:related} states how existing systems differ in where the assembly is carried. Section~\ref{sec:model} states the conditions and the advantages they carry, Section~\ref{sec:construction} builds the SDK, and Section~\ref{sec:framework} fixes the rules of the comparison. Section~\ref{sec:results} shows that the session survives a fault of its infrastructure, that the capability set changes while the system runs, and what the advantages cost, and Section~\ref{sec:conclusion} concludes. The supplementary materials give the formal support and the proofs, read the construction part by part, and report the experimental environment and the secondary trials.

\section{Background and Related Work}
\label{sec:related}

\begin{figure}
\centering
\includegraphics[width=\linewidth]{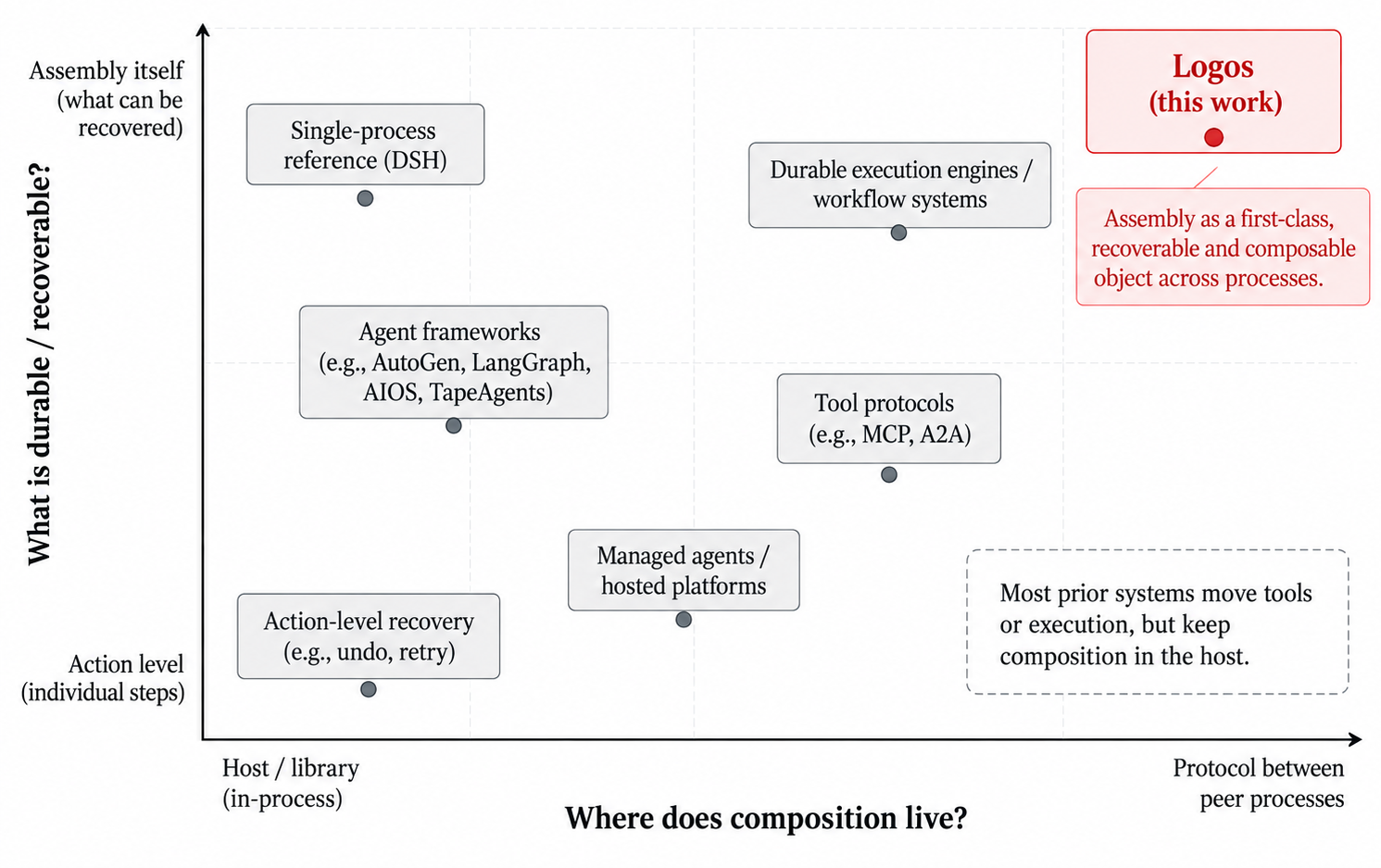}
\caption{Where composition lives across the surveyed systems, from in-process libraries to protocols between peer processes. Logos alone places the assembly itself in the protocol and keeps it recoverable.}
\label{fig:landscape}
\end{figure}

This section first gives the background the conditions build on and then the systems directly related to this work. Existing systems divide by where composition and recovery reside, some keeping composition as library objects of the host process, some using protocols that move tools or execution out while composition stays with the host, and a further group keeping durable records whose recovery remains inside one process, one library, or one platform, a group that includes a single-process realization of the calculus that stands behind this work. Table~\ref{tab:whatmoves} compares the frameworks by what each moves out of the process, and Figure~\ref{fig:landscape} places them by where composition lives.

\begin{table}[!b]
\caption{What each framework moves out of the process, and what stays in it.}
\label{tab:whatmoves}
\begin{tabular}{@{}p{0.19\linewidth}p{0.35\linewidth}p{0.34\linewidth}@{}}
\toprule
Framework & Moves out of the process & Stays in the process \\
\midrule
MCP & tool servers and calls & composition and session state \\
A2A & task delegation among agents & internal assembly of every agent \\
AutoGen, LangGraph, AIOS, TapeAgents & nothing & all capabilities as in-process objects \\
Codex CLI & nothing & session replay inside one process \\
Managed Agents & hands, the sandboxes and tools & the brain and its assembly, as a hosted service \\
Durable execution engines & execution & assembly and the workflow definition \\
OpenHands & sandboxed execution & composition in the agent host \\
DSH (DeepSeek Harness) & nothing & the whole harness and every record in one process \\
Logos & composition and assembly, state into the transcript & nothing but a rebuildable routing table \\
\bottomrule
\end{tabular}
\end{table}

\subsection{In-Process Composition}

The systems of this family keep composition as library objects of the host process. The survey by Li et al.~\cite{li2026agentharness} maps more than 170 open-source projects onto seven layers, and single-loop coding agents~\cite{yang2024sweagent}, orchestration frameworks~\cite{wu2024autogen,hong2024metagpt,kim2024llmcompiler}, and pipeline runners organize capabilities the same way. Assembly is not a runtime operation, no record of mounting exists, nothing can be unloaded or reverted, and the host process is the failure domain (Sections~\ref{sec:failure} and~\ref{sec:response}). DSPy compiles a pipeline of modules and tunes the prompts and demonstrations of each module as the parameters of the program~\cite{khattab2024dspy}.

\subsection{Protocols That Move Tools Out}

The protocols of this family move tools or conversations out of the process, and the composition of the agent stays in the host. The nearest is the Model Context Protocol, which moves tool servers and calls out of the process through a host, a client, and a server, exchanging tools, resources, and prompts as typed JSON-RPC messages~\cite{hou2026mcp,ehtesham2025}, the same decomposition microservices make for ordinary services~\cite{accetta1986mach,dragoni2017microservices}, and its specification defines initialization, operation, and shutdown, with no mount, no unmount, and no reversal among its words. A second line standardizes conversation among agents, the Agent-to-Agent protocol~\cite{acp2025,christie2025argus} and the Agent Communication Protocol~\cite{ehtesham2025}, neither specifying the internal composition of an agent. OpenHands separates agent logic from sandboxed execution, composition staying with the agent host~\cite{wang2025openhands,felter2015containers}. In all of these, composition and session state remain in the host, Logos has no host process in which they could remain, and none carries mount, unmount, or reversal semantics.

\subsection{Durable State and Its Carrier}

The systems of this family keep durable records, and the recovery of those records stays inside one process, one library, or one platform. LangGraph~\cite{langgraph2024} orchestrates agents as graphs of stateful nodes with checkpointed state, resuming an interrupted run from the last checkpoint. AIOS~\cite{mei2025aios} isolates scheduling, context, and memory management into an agent kernel, MemGPT manages the context of one agent the way an operating system manages memory~\cite{packer2023memgpt}, and Codex CLI~\cite{openai2025codex} reconstructs a session by replaying a recorded file inside one process~\cite{li2026agentharness}. TapeAgents records the whole session as a granular, structured, append-only tape that is also the session's resumable state, the record closest in form to the transcript of Logos~\cite{bahdanau2025tapeagents}, and in each of these the persistent record restores the state of one computation, inside the process or library that wrote it.

Cognitive architectures place memory among the core modules of a language agent~\cite{sumers2024coala}, Mem0 runs a two-phase extraction and update pipeline over a conversation~\cite{chhikara2025mem0}, and Zep maintains memory as a temporal knowledge graph whose facts carry validity intervals~\cite{rasmussen2025zep}, records of knowledge in which assembly and recorded inverses have no place.

The member of this family closest to this work is Temporal~\cite{temporal2025docs}, which persists every step of a workflow in an append-only event history and recovers by replay, the event-sourcing form~\cite{fowler2005eventsourcing}, with deterministic workflow code, while DBOS holds workflow state in a database~\cite{kraft2021dbos}, AWS Step Functions in a declarative state machine~\cite{awssfn2025}, and Azure Durable Functions in orchestrator functions the platform replays~\cite{azuredurable2025}. Determinism gives Temporal exact replay, while after a cold swap the model may answer differently (Section~\ref{sec:scope}), and the transcript replay then checks four invariants on every reconstruction (Section~\ref{sec:transcript}).

The hosted architecture closest to this work is Managed Agents, which decouples the brain, the hands, and the session, the session an append-only event log, and a failed brain reboots through wake(sessionId) and resumes from the last recorded event~\cite{anthropic2026managed}. The record and the restart take the same form in Logos. The recovery machinery lives inside one operator's infrastructure, itself a failure domain, while in Logos no process owns recovery (Section~\ref{sec:response}). Managed Agents recovers the session of one brain, mounting, unmounting, and reverting not among its operations, while Logos recovers the assembly itself under the stated conditions (Section~\ref{sec:model}). The survey names as an open problem the reconstruction of missing agent state from durable artefacts rather than reliance on compressed history~\cite{li2026agentharness}, and the transcript with its replay invariants is that reconstruction.

\subsection{The Single-Process Reference and Action-Level Undo}

DeepSeek Harness realizes the spatiotemporal-composability calculus as a harness in which every capability is a plugin inside one Node.js process, and its session log is an in-process append-only record~\cite{dsh2026spatiotemporal,dsh2026repo}, so plugins, sessions, and recovery records share one physical failure domain. The conditions lift the guarantee to an arbitrary assignment of components and records to processes (Section~\ref{sec:generalization}), and under the fault of the evaluation the reference lost every faulted session (Section~\ref{sec:taskcmp}).

A separate line of work makes the actions of an agent reversible, compensation as a disposition for the irreversible in sagas~\cite{garcia1987sagas}, GoEX managing checkpoints and undo inside a single runtime~\cite{patil2024goex}, ReAgent making each reasoning step reversible~\cite{zhao2025reagent}, AgentRewind rolling a long-horizon execution back to aligned checkpoints~\cite{zhuang2026agentrewind}, and the Revisable by Design theory grading every action as idempotent, reversible, compensable, or irreversible~\cite{zhai2026revisable}. In this line the unit of reversibility is the action or the checkpoint, and the recovery record stays inside the runtime that wrote it, while in Logos the unit is the assembly and the carrier a transcript that persists beyond the runtime. Atomix settles tool effects transactionally behind adapters~\cite{mohammadi2026atomix}, and Concordia keeps the device state of long-running inference recoverable on a persistent GPU kernel~\cite{gan2026concordia}, each covering one computation. Two preprints study adjacent questions, AgentGit~\cite{li2025agentgit} and $\lambda_A$~\cite{liu2026lambdaa}, and neither places assembly on a protocol or examines recovery across processes.

\subsection{Key Differences}

Over the systems surveyed above, this work adds three things. The contribution of this paper is the conditions, Logos is one SDK that fulfills them, and the measurements of Section~\ref{sec:results} are the evidence of fulfillment. No surveyed system carries composition and its record out of the host, and Logos is the only one that does, with mounting, unmounting, and reverting as protocol actions between peer processes under stated conditions, measured and fault-tested. None of the surveyed systems states the conditions under which a recovery guarantee survives the crossing of process boundaries, and none reports a failure-domain measurement of composition under injected kills. The conditions, the bus protocol, and the measurements supply these three (Sections~\ref{sec:generalization}, \ref{sec:response}, and~\ref{sec:changes}).

\section{Conditions of Host-Free Composition}
\label{sec:model}
\label{sec:generalization}

The conditions of this paper turn a plugin-based agent into a construction that needs no host process.

A session advances in rounds over a transcript $T$ which depends on no central host. The engine appends the result of a tool call to the transcript first, and only then settles the round and broadcasts the new entry to every node~\cite{eugster2003publish}. Formally, at round $t$ the LLM execution engine runs
\begin{equation}
\label{eq:loop}
x_t = \omega(\pi(T_t)), \qquad a_t = \chi(x_t), \qquad T_{t+1} = T_t \cdot \rho(a_t),
\end{equation}
where $\pi$ projects the transcript to the message history, $\omega$ is input synthesis, $\chi$ is the stateless model call, and $\rho$ is output settlement.

In the calculus~\cite{dsh2026spatiotemporal}, a capability is a component that carries the recorded means of undoing its effect, the tracked inverse, and a fault is the death of a process, which removes every component and record resident in it. This paper therefore relaxes co-residence to an arbitrary assignment $\alpha$ of components and records to $N$ processes. To state faithfulness, this paper writes $\tau$ for an abstract trace of assembly and removal steps in the state space and $\Pi$ for the macrostate projection, the view that keeps all that an observer can distinguish. The assignment is faithful when for every abstract trace $\tau$ there is an implemented trace $\tau'$ with
\begin{equation}
\label{eq:faithful}
\Pi(\tau')=\Pi(\tau).
\end{equation}

\subsection{The Conditions}
\label{sec:conditions}

The conditions rest on two modelling facts, that the model call is a stateless pure map $y=\chi(x)$, and that the synthesis $\omega$ and the settlement $\rho$ run outside $\chi$, which place the cross-round state outside the model call. Four engineering conditions turn them into a guarantee that distributed records recover faithfully, stated in the notation of the supplementary materials, with $p(c)$ the process of $c$ under $\alpha$,
\begin{equation}
\label{eq:conditions}
\begin{gathered}
\text{C1 (externality): } x=\omega(S),\ S \text{ outside every model call},\\
\text{C2 (carrier): } \rho_{C}(a_i)\simeq g_i\ \forall i\ \Rightarrow\ \Phi_{C}(T)\simeq\Phi_{\mathcal{D}},\\
\text{C3 (localization): } c_m\perp c_n\ \Rightarrow\ \mathrm{rec}(c_m)\subseteq p(c_m),\\
\text{C4 (resolution): } \mathrm{resolve}(c,t)=R_t(c),\ R_t=\mathrm{fold}(e_1\cdots e_t).
\end{gathered}
\end{equation}

C1 handles state placement. The cross-round state of the session lies outside the model call, so any process can read the shared carrier and prepare the next input.

C2 handles the record carrier. The transcript holds every tracked inverse, and the recovery a replacement process reads from the record is observationally equivalent to the recovery the original process would have executed in place.

C3 handles recovery coordination. The recovery data of a component is held only by the process the component lives in, so mutually independent components do not re-form a shared failure domain through a shared recovery context.

C4 handles provider resolution. The search for a provider reduces to one routing-table lookup folded from the broadcast record, and every process that has seen the same broadcast sequence reaches the same table.

Every recovery equality in the calculus holds up to an observational equivalence, and the symbol $\simeq$ of C2 names it. Two states are observationally equivalent when no observer of the context can tell them apart, and the parts of the context that an observer can read are called its coeffects.

\subsection{The Supporting Disciplines}
\label{sec:disciplines}

\begin{figure}
\centering
\includegraphics[width=\linewidth]{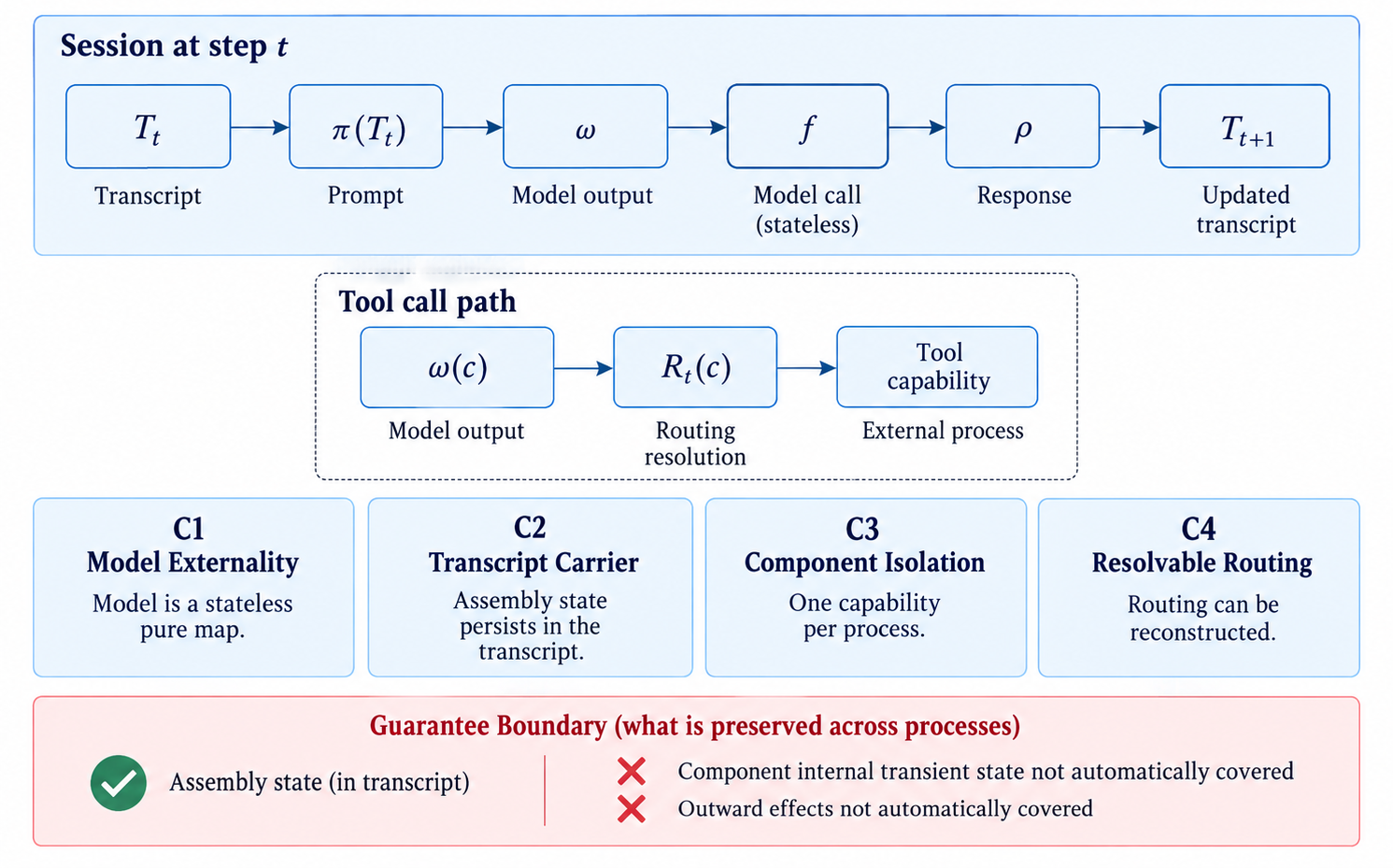}
\caption{The session loop over the transcript. The tool-call path runs through the routing table. The four conditions are C1 through C4. The guarantee boundary marks what crosses processes and what stays outside.}
\label{fig:model}
\end{figure}

The conditions hold together with three disciplines. Every revertible effect records its anchor, the entry that pins the effect and its inverse, to a persistent store at the step of occurrence, the store condition. Every ID admits exactly one writer, the writer condition. The delivery discipline, the rules by which calls, responses, and notifications reach the processes, carries three constraints. E1 pairs every response with its own call, so the transcript records every abstract step. E2 gives each node id one registration, with an explicit refusal for a conflicting claim, so a replacement process can inherit its predecessor's place. E3 gives every observer the same sequence of supply-change notifications, so every node folds the same table. Figure~\ref{fig:model} summarizes the session loop, the tool-call path, the four conditions, and the guarantee boundary.

\subsection{The Agent the Conditions Construct}
\label{sec:constructed}

Under the conditions and the disciplines, an implementation over $N$ processes is a faithful implementation of the calculus, Eq.~\eqref{eq:faithful} holds for an arbitrary assignment of components and records to processes, and the formal statement with its proof is one of the supports of this paper, the supplementary materials (S.1), which also proves two of the disciplines necessary (S.2). The construction reads in both directions. Meeting the conditions constructs a plugin-based agent with no host process, and removing any one part that discharges them makes the assembly need a host again (the supplementary materials, S.3).

\begin{figure}
\centering
\includegraphics[width=\linewidth]{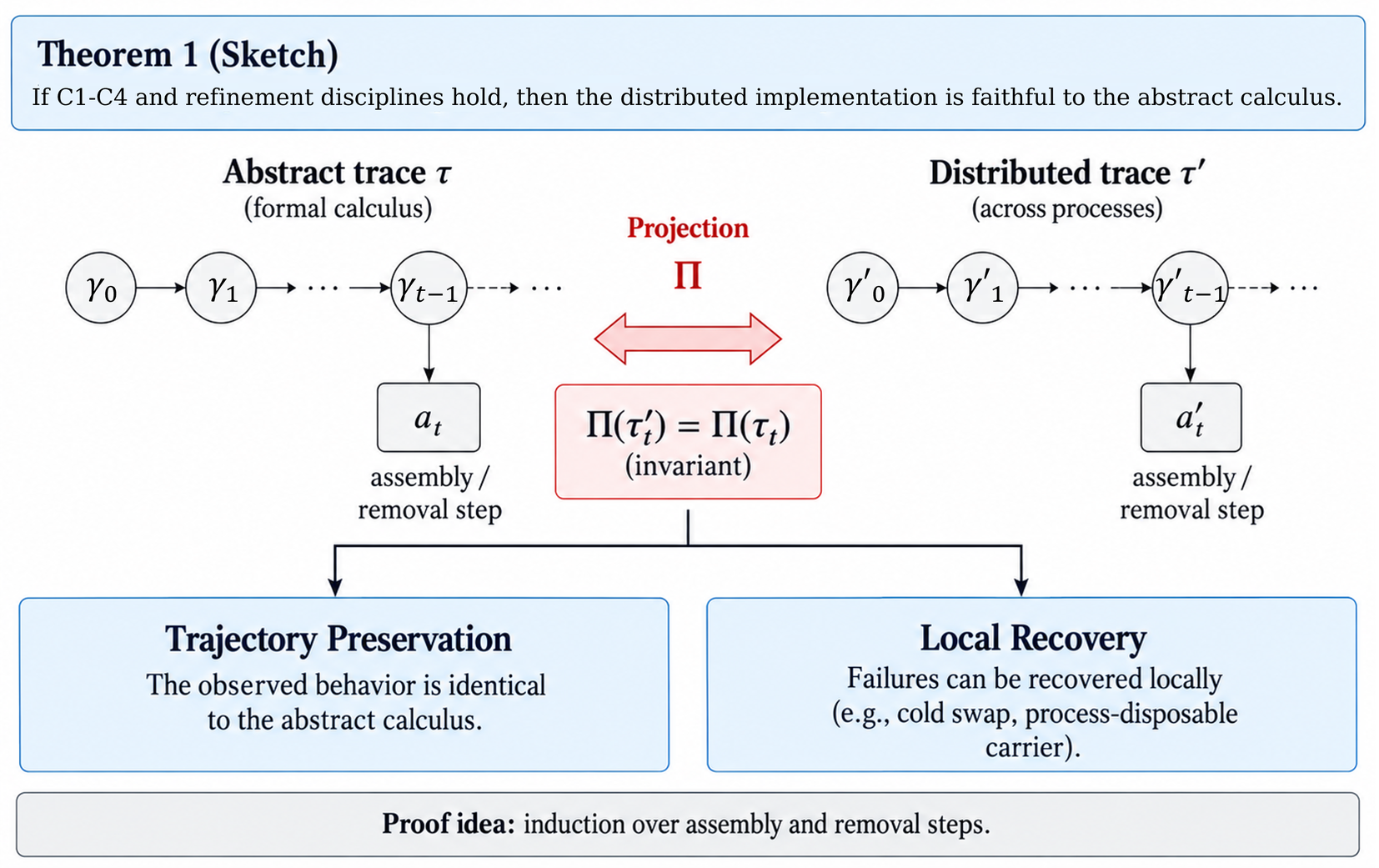}
\caption{The formal support in one picture. The abstract trace and the distributed trace meet under the projection. The two consequences are trajectory preservation and local recovery.}
\label{fig:trace}
\end{figure}

Two properties of the support carry the meaning. The implemented trace and the abstract trace carry the same macrostate at every step, a state determined by the anchor alone, so any process that reads the transcript prefix reaches the same macrostate, and the semantics holds over $N$ processes as it held over one. The undo accumulator, the running composition of the recorded inverses, is rebuilt from the transcript by a replacement process, and the inverse data of a component stay in the process the component lives in, so the undo capability crosses the process boundary, and a process becomes a carrier that can die and be replaced without losing it. Figure~\ref{fig:trace} sketches the two properties for the traces.

An agent constructed this way carries five advantages, and the rest of this paper measures each of them. No process holds the privileged position of a host, so the death of any process removes only the execution running in that process, and the session survives (Sections~\ref{sec:taskcmp} and~\ref{sec:missions}). A fault costs no wasted calls and no lost records, the explicit refusal closes the dead endpoint, and the ownerless transcript keeps every record (Section~\ref{sec:taskcmp}). Recovery is local to the faulted process, a replacement rebuilding from the record with no cross-process coordination (Sections~\ref{sec:response} and~\ref{sec:guarantee}). The capability set grows, shrinks, and replaces its members while the system runs, mounting, unmounting, and replacement as protocol actions (Section~\ref{sec:changes}). The undo semantics of the calculus carries across processes, the tracked inverse rebuilt from the transcript and replayed after a cold swap (Section~\ref{sec:guarantee}).

\subsection{The Guarantee Boundary}
\label{sec:boundary}

The guarantee covers only an effect whose location the system can modify exclusively and restore to the state before the modification. When both hold, the effect lies inside the guarantee boundary, and when either fails, it lies outside, untracked by the calculus and not redone by the recovery mechanism of this paper.

Some operations cross this boundary midway through their execution, a payment to the external world for example, which the system can record but cannot restore. For such an effect the system has two dispositions, withholding, deferring the emission until the state that produced it has been made durable, and compensation, which restores not the original state but a coarser observational equivalence class, in which more states count as the same.

C3 presumes the pairwise independence of component effects, formally Definition 3 of the supplementary materials. Effects whose order changes the result are either serialized through single-writer tools or placed outside the boundary of Definition 7 of the supplementary materials. The presumption holds in this system, because the shared mutable state is a table whose operations on distinct entries commute~\cite{dsh2026spatiotemporal,shapiro2011crdt}.

\subsection{Deployment Location}
\label{sec:deployment}

Neither the protocol nor the routing table records the machine a process runs on. Nodes consume capabilities by name, two machines each carry their own bus joined by a network, and each is one node on the other's bus, under the same registration and the same forwarding path.

The framework also covers the case of disconnection. A node whose internal state changes while disconnected can no longer be rebuilt from the record. When the node regains its connection it mounts as a new node, and whatever its outage left unrecoverable falls outside the guarantee boundary and receives one of the two dispositions of Section~\ref{sec:boundary}.

\section{The Cross-Process Composition Framework}
\label{sec:construction}

\subsection{System Design}
\label{sec:design}

\begin{figure}
\centering
\includegraphics[width=\linewidth]{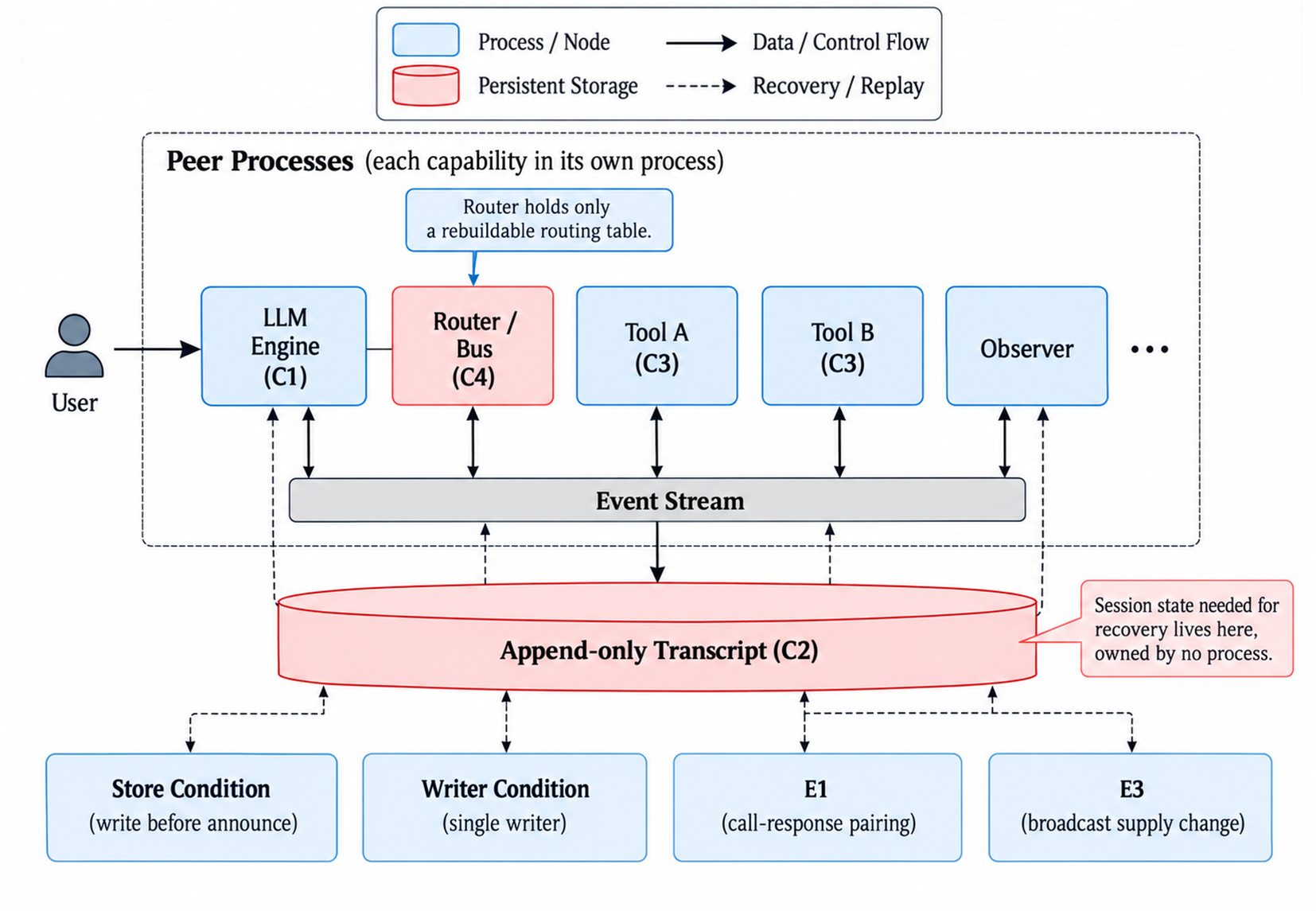}
\caption{The framework mapped to its formal support. Peer processes run on the bus. The router holds only a rebuildable routing table. The event stream and the append-only transcript carry the session state needed for recovery.}
\label{fig:arch}
\end{figure}

The conditions of Section~\ref{sec:model} leave the assignment of components and records to processes free, and this paper fixes one peer-process assignment, the paradigm of the Robot Operating System~\cite{quigley2009ros,macenski2022ros2}, realized as Logos. The framework makes two decisions first. The capability assembly is placed on a bus shared by peer processes, and the session record is kept in an append-only carrier that no process exclusively owns. Figure~\ref{fig:arch} shows these parts and their relations.

\subsection{The Session Record}
\label{sec:transcript}

The transcript is an append-only JSON Lines (JSONL) file owned by no process, and every step of every session is written to it at the moment it occurs~\cite{chandy1985snapshot}, so the death of a process does not take away what has been written. The file allows exactly one update,
\begin{equation}
\label{eq:appendonly}
T' = T \cdot e, \qquad \mathrm{idx}(e) > \max_{e' \in T}\, \mathrm{idx}(e'),
\end{equation}
an append with a fresh identifier, no deletion, and no rewrite.

The model never reads the file directly. Messages shown to it are a projection $\pi: E^{*} \to M^{*}$, long tool results pruned in the projection while the full text remains in the file. The replay of a transcript is checked against four invariants. I1 requires every entry to lie inside one pair of a turn start marker and its matching turn end marker. I2 requires identifiers to be monotone~\cite{lamport1978time}. I3 pairs every tool use $u$ with exactly one result $r$ of the same id, and I4 requires the rebuilt projection to equal the messages the session showed, $\pi(T) = M$. The settlement order is durable before visible,
\begin{equation}
\label{eq:settle}
\mathrm{write}(e) \prec \mathrm{announce}(e),
\end{equation}
the store condition enforced at the moment the step occurs.

\subsection{The Bus}
\label{sec:bus}

The router is a single Go process, its three duties registration, forwarding, and broadcast. It schedules nothing and reads no business content, and the bus is itself carried by an ordinary killable process, replaceable under the same id, so no node depends on it as a host (Section~\ref{sec:response}). The wire protocol is newline-delimited JSON (NDJSON) over TCP, one object per line, with frames up to 64 MB.

Every node holds one connection to the router and every message passes through it, each connection a bounded outbound queue with control messages paired by global call ids and never dropped, the pairing discipline of concurrent objects~\cite{herlihy1990linearizability}. The delivery rules are written as relations over one connection queue with drop set $D$, control class $C$, and lossy stream $L$.
\begin{equation}
\label{eq:delivery}
\begin{gathered}
|Q| \le 1024, \qquad D \cap C = \emptyset,\\
e_j \in D \wedge i < j \,\Rightarrow\, e_i \in D, \qquad
e \in L \wedge \mathrm{age}(e) > 5\,\mathrm{s} \Rightarrow e \in D,
\end{gathered}
\end{equation}
A lossy drop is prefix-closed in backlog order, so it never reorders, and beyond these rules the protocol demands nothing further, so any process that implements them can join as a node, whatever language it is written in, the implementation constructing peer nodes in Go, Python, and Node.js on one bus.

\subsection{The Routing Table}
\label{sec:routing}

The routing table maps a node id to a connection with its role and provided capabilities. Any node can read the whole table at any time, registered or not, and no process holds the only valid copy. The events are $\mathtt{online}(n,c)$, $\mathtt{offline}(n,c)$, and $\mathtt{leaving}(n,c)$ over node ids $\mathcal{N}$ and capability names $\mathcal{C}$, the supply relation $P_t \subseteq \mathcal{N} \times \mathcal{C}$ folds them by applying the online pair and removing it otherwise, the table $R_t(c)$ is its projection onto capability names, and resolution is the lookup $\mathrm{resolve}(c,t) = R_t(c)$. Any process that reads the whole broadcast record from the start folds the same table, so no process needs to maintain it.

The fold can return several providers for one capability. The table lists every provider of a capability without distinction, and the querying side selects one provider under a fixed rule. Each node consumes capabilities by name and never by node identity, and the dependency of node $n$ on capability $c$ is in one state, waiting, active, or closed.

Under E3 every node observes one event sequence, so the fold is deterministic and four properties hold. The dependency state is a function of the table and the log alone, so resolution survives process death. When one provider of $c$ departs and another arrives, the dependency rebinds from active to active and no consumer restarts. Every node holds the same table, $R_t^{(n_1)} = R_t^{(n_2)} = R_t$. When $R_t(c)$ is empty, an in-flight call receives an explicit error naming $c$, and the return of a provider costs one re-resolution and no restart.

One node id admits one row, and a conflicting claim receives an explicit refusal, the writer condition enforced at registration.

\subsection{Nodes}
\label{sec:nodes}

A node is any process that follows the protocol, and it mounts as a peer, uniformly, as in the actor model~\cite{hewitt1973actor,agha1986actors}. The framework registers its nodes in three roles, LLM execution engine, tool, and observer, the observer receiving every broadcast and announcing nothing. An LLM execution engine runs the loop of Eq.~\eqref{eq:loop} over one session, and a tool provides a capability. Algorithm~\ref{alg:harness} gives the loop.

\begin{algorithm}[t]
\caption{LLM execution engine loop with recovery}
\label{alg:harness}
\begin{algorithmic}[1]
\REQUIRE session id, transcript
\STATE import the transcript, rebuild the projection
\LOOP
\STATE synthesize input from the projection
\STATE call the model, stream the answer into the transcript
\IF{the answer requests a tool}
\STATE route the call, append the result to the transcript
\ENDIF
\STATE settle outputs, broadcast the change
\STATE \COMMENT{when the process running the session dies, its replacement resumes from line 1}
\ENDLOOP
\end{algorithmic}
\end{algorithm}

\subsection{Component Lifecycle}
\label{sec:lifecycle}

A capability is a process. Tools mount at their own endpoints, and the router stays unchanged throughout. Mounting with a missing provider leaves the dependency state in waiting, the dependent's calls answered by an explicit refusal naming the missing capability, and a later online broadcast activates the dependency without code change on any side. A provider going offline triggers re-resolution, a replacement provider of the same capability rebinds the dependents, and a provider announcing departure waits for its confirmed dependents to leave before it withdraws. The discipline is a soft layer over the late binding of the bus, resolution by name at call time, an agreement the nodes follow that the bus itself does not enforce. The order follows the calculus, supply preceding consumption and withdrawal following consumption.

\subsection{Failure Handling}
\label{sec:failure}

In the single-process reference the death of the hosting process restarts the whole stack. On this bus the death of the process running a session costs that process alone~\cite{armstrong2003thesis}. When the process running a session dies, a new process $P'$ of the same id imports the transcript, rebuilds the projection, and continues the session, a cold swap. Writing $T_\kappa$ for the transcript at the kill point $\kappa$, recovery satisfies
\begin{equation}
\label{eq:coldswapexec}
\begin{gathered}
\mathrm{view}(P') = \pi(T_\kappa) = \mathrm{view}(P_\kappa),\\
\mathrm{redo}(P') = \mathrm{acts}(T) \setminus \mathrm{recorded}(T_\kappa),
\end{gathered}
\end{equation}
Recovery performs no rollback, and the calls redone are exactly those the transcript had not recorded.

The death of the router is a separate case. The router holds only its routing table, so its death suspends only the routing of new messages, model streaming and session processes continuing uninterrupted, and in-flight bus calls waiting for its return.

After a cold swap the recovering process reissues the calls the transcript had not recorded, so repetition must be safe, and three classes of effect are admissible on the bus, the replayable effects, whose repetition is observationally neutral, covering the benchmark tools under the scoping of Section~\ref{sec:framework}, the idempotent effects, keyed so that a repeated call settles to the same result, and the irreversible outward effects, which stay outside the boundary of Definition 7 of the supplementary materials and are handled by withholding or compensation, all three grounded in constructs of the calculus~\cite{dsh2026spatiotemporal}.

\subsection{The Framework Mapping}
\label{sec:mapping}

The framework realizes the conditions of Section~\ref{sec:conditions} construct by construct, the correspondences stating four obligations, one per condition of Eq.~\eqref{eq:conditions}, each discharged by named parts.

\textit{State placement, C1.} The runtime of the framework is exactly the pair $(\omega,\rho)$ of Eq.~\eqref{eq:loop}, the shared state $S$ of C1 is carried by the transcript projection $\pi(T)$, and the model registers as an ordinary dispatchable node, so no component requires co-residence with $\chi$.

\textit{Carrier, C2 with Corollaries 1 through 3 of the supplementary materials.} The session record is the append-only transcript of Section~\ref{sec:transcript}, the free-monoid carrier of Corollary 2, with no owning process, and a dead session process is replaced by cold swapping through the composite accumulator of Corollary 3, measured in Section~\ref{sec:guarantee}.

\textit{Recovery, C3.} A capability is a process $p(c)$, the inverse data $\mathrm{rec}(c)$ reside in $p(c)$ alone, and the lifecycle of Section~\ref{sec:lifecycle} re-resolves dependencies with no cross-process coordination of recoveries.

\textit{Resolution, C4.} The router holds only the table $R_t$ folded from the broadcast record, the single-writer condition holds at registration, and under E3 every observer folds the same sequence into the same $R_t$.

Table~\ref{tab:correspondence} indexes the mapping row by row. The supplementary materials read the table from the removal side, part by part, and removing any one of seven of the eleven parts makes the assembly need a host again. The framework meets E1 through the NDJSON protocol and the bounded queues, E2 through the registration conflict refusal, E3 through per-observer delivery queues, the store condition at settlement, and the writer condition at registration, all audited in Section~\ref{sec:premises}.

\begin{table}[!t]
\caption{The framework mapping. Each row puts one part of the Logos framework against the calculus construct it realizes and the condition it satisfies.}
\label{tab:correspondence}
\begin{tabular}{@{}p{0.36\linewidth}p{0.30\linewidth}p{0.22\linewidth}@{}}
\toprule
Construction part & Calculus construct & Condition \\
\midrule
a capability is an independent process & the component of Definition 1 & recovery localization \\
the model is an ordinary dispatchable bus node & the stateless pure map $\chi$ & orchestration externality \\
the transcript is append-only and plaintext & the free monoid $\mathrm{Act}^*$ on the generators of the inverse data & carrier substitution \\
cold swapping resumes the same id & the composite accumulator $\Phi(T)$ & Corollary 3 \\
the router routes and registers only, holding no business state & $F_\gamma$ and the provider map read by ID & external resolution \\
the registration refusal of a duplicate id & the single writer of a well-formed registry & the writer condition \\
the online and offline broadcasts & the change classification $\mathrm{notify}_d$ & E3 \\
the broadcast order over per-observer queues & supply before demand with withdrawal after consumption & E3 \\
the routing in use, the current resolution & the committed view & Lemma 4 \\
outward effects stay outside the boundary & Definition 7 & the boundary of the calculus \\
every event is broadcast and replayable & the computable projection $\Pi$ & Definition 5 \\
\bottomrule
\end{tabular}
\end{table}

A deployment as peer processes is admissible under two conditions. The bus time scale must separate from the model time scale, a millisecond hop against a first token nearly three orders of magnitude slower, and the topology must be stationary at the model's step scale, nodes neither joining nor leaving between one model step and the next.

\subsection{The Framework at Work}
\label{sec:operation}
\label{sec:case}

\begin{figure}
\centering
\includegraphics[width=\linewidth]{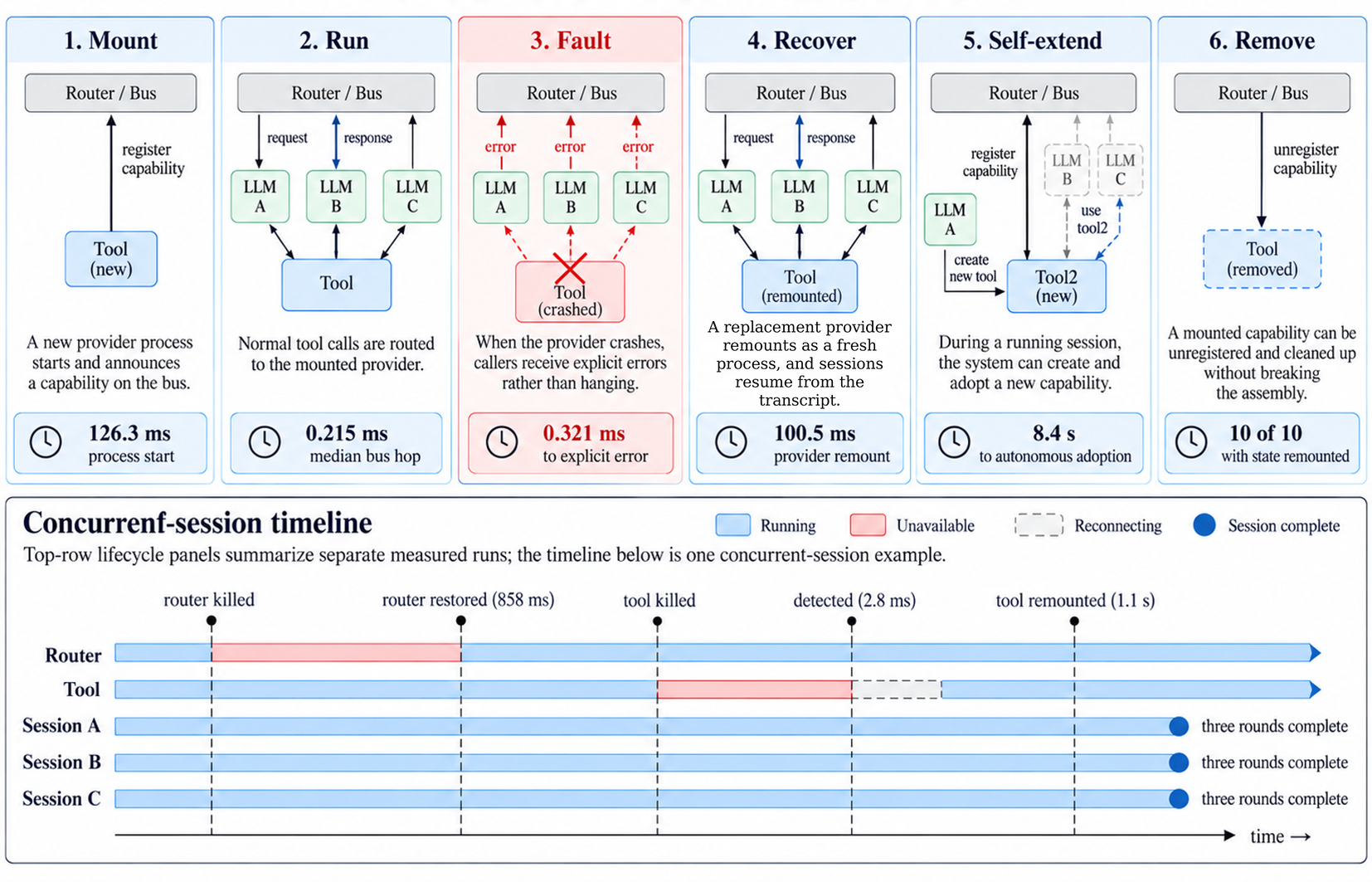}
\caption{The six stages of a capability on the bus, with the measurement of each stage. The timeline below follows the concurrent sessions through two injected faults.}
\label{fig:lifecycle}
\end{figure}

At run time the parts form one timeline, and Figure~\ref{fig:lifecycle} follows one capability from mounting to removal. Each tool registers, the registration becomes an online broadcast, and every node folds the event into its table. At the end of startup the assembly exists in the tables and the append-only files, and no process holds it.

The capability set changes through the same broadcasts. A mounted tool resolves with no prompt naming it. An unmounted tool goes offline, and a call that still names the capability receives an explicit refusal. A replacement exchanges the provider process under the same name and the same schema, and the dependents rebind with no consumer restart.

A fault is the death of one process, and each kind of death meets its own mechanism. A dead tool goes offline, the next call receives an explicit refusal, and the external watchdog restarts the tool from a blank world. A dead session process is detected outside the system, and a new process of the same id imports the transcript and continues the session. In each case the assembly outlives the process that died, and Sections~\ref{sec:framework} and~\ref{sec:results} test this property.

\section{Experimental Evaluation}
\label{sec:framework}
\label{sec:results}
\label{sec:resultsguarantee}

This section fixes the rules of the comparison first and reports the measurements after. The rules cover the configurations, the tasks, the fault, and the measured quantities, and the subsections that follow report each measurement in turn.

\subsection{Setup}
\label{sec:setup}

\subsubsection{Configurations}
\label{sec:configs}

The evaluation compares three configurations under one fault action, and the configurations differ only in how the agent is arranged across processes. In Logos the assembly is on the bus, the session state is in the transcript, and no host process exists. In the MCP configuration, tool servers run as external processes over the official SDK, but the assembly and the session state stay inside the host. In the single-process reference DSH (DeepSeek Harness), the same session code is composed from the reference implementation of the calculus~\cite{dsh2026repo}, version v0.1.0-rc.5, with every plugin, session, and record inside one process.

The two baselines bound the host paradigm from both sides, the single-process form and the cross-process form in industrial use. Everything else is held fixed, the tasks, the model and its endpoint, the tool set, the system prompt, and the fault action, and no configuration receives tuning beyond the form in which it ships.

\subsubsection{Tasks, Fault, and the Say-Once Rule}
\label{sec:tasks}

The tasks are the full 200 of the Berkeley Function Calling Leaderboard (BFCL) v4 multi-turn base~\cite{patil2025bfcl,patil2023gorilla}, run under DeepSeek V4 Flash and mimo-v2.5-pro. A task is a scripted conversation of several user turns, and the agent answers each turn, calling a tool where the instruction needs one. The tools keep a working state for the task, the task world, judged against the ground-truth final state.

Each task was assigned one scheduled fault opportunity, a kill of the tool process at the first tool call of the second round. The fault was realized in 197 of 200 runs under DeepSeek V4 Flash for each configuration, and under mimo-v2.5-pro in a different number of runs in each configuration, so the evaluation distinguishes all-task outcomes from mechanism results conditioned on a realized fault. A 60-second watchdog restarts the tool with a blank world, and the engine proceeds with its remaining turns. The kill arrives from outside the system, each task runs once, and sampling follows the default settings of the model endpoints.

The action lands on a tool node in Logos, on a tool server outside the host in the MCP configuration, and on the tool layer inside the single process in the reference. One shared rule for recovery, say-once, makes the comparison fair. The instruction of each round is delivered exactly once, and recovery proceeds without re-sending any round. If rounds could be delivered again after a fault, no configuration would lose a session and the survival comparison would measure no difference. The tools of these measurements act inside the task world alone, so a call repeated after recovery acts only on the world the restart resets.

\begin{figure}
\centering
\includegraphics[width=\linewidth]{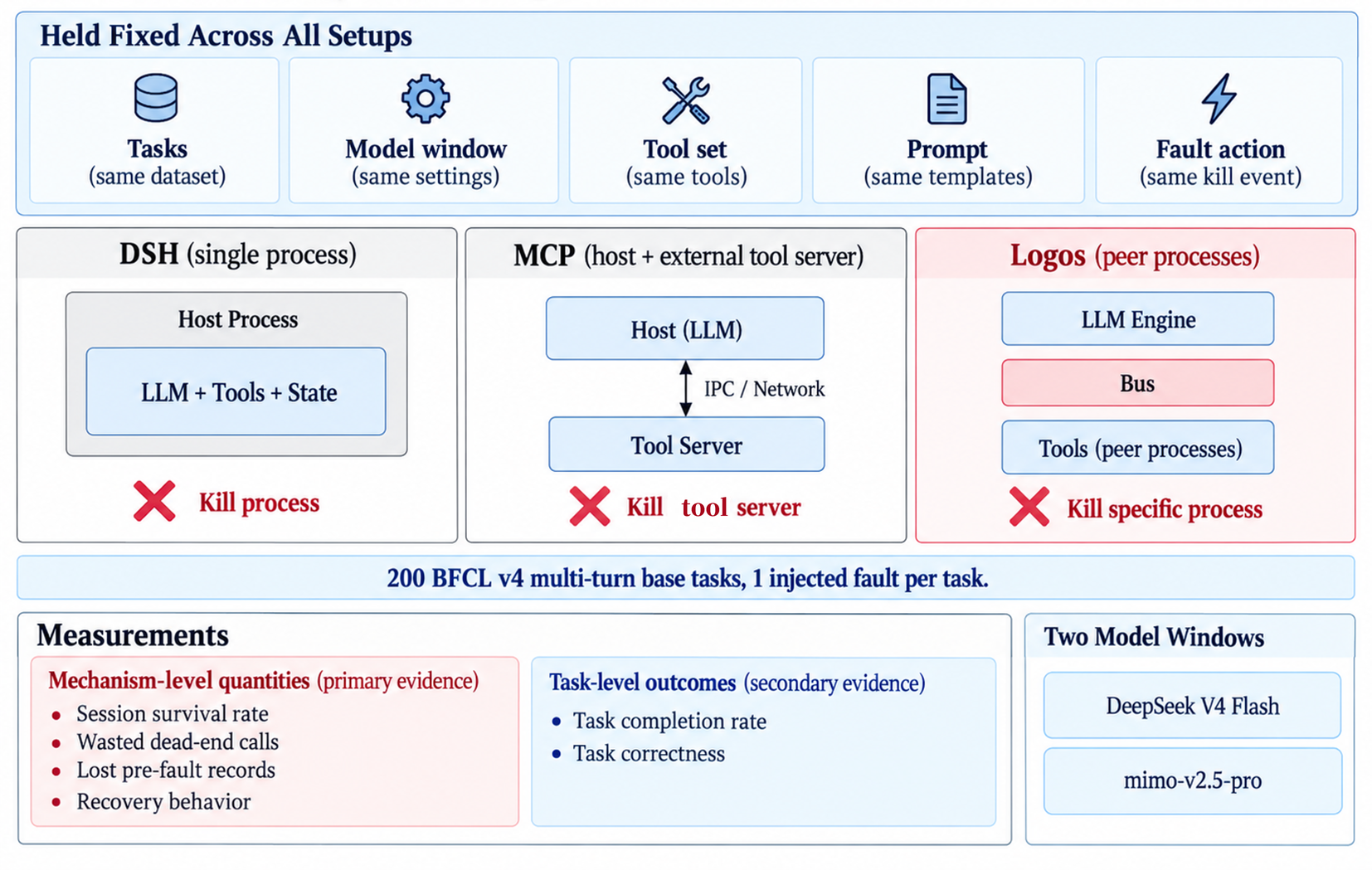}
\caption{The controlled comparison. The variables held fixed across the three configurations are listed first. The injected fault lands at a different point in each configuration. The measurements are mechanism-level quantities as primary evidence and task-level outcomes as corroboration.}
\label{fig:framework}
\end{figure}

\subsubsection{Predictions and Measured Quantities}
\label{sec:predictions}

Three invariants serve as predictions of the conditions, zero wasted calls, zero lost records, and the survival of every faulted session, called the identities. Through the explicit refusal, the ownerless transcript, and the cold swap, the advantages become observable in the running system. A fault ends at the process that carries it, no call crosses to a dead endpoint, and no record dies with a process. The refusal may also be emulated from outside, and Section~\ref{sec:hold} measures that emulation.

A session counts as surviving the fault when its record stays continuous and a process continues the session from that record. The reference reopens the whole session, its tool layer unreachable, and because the say-once rule forbids re-sending any round, a reopened session is a new session, so the faulted sessions of the reference count as lost. The official validator replays the whole trajectory into a fresh instance of the task world, and a task succeeds when the ground-truth final state is contained in the final state the run produces~\cite{patil2025bfcl}.

The SDK is measured on seven quantities, namely detection time, restoration time, outage duration, calls to the unavailable endpoint, lost records, redone calls, and bus overhead relative to model latency. The first three characterize recovery time, the next three recovery cost, and the last tests whether the distributed form adds material latency relative to the model call. The mechanism-level quantities are primary, taken under both task models on the same 200 assigned tasks, with session survival reported against each model's realized-fault counts. Task success and the official-validator score are secondary end-to-end outcomes, and they corroborate the mechanism measurements.

Every assertion is checked against two separately recorded on-disk audit views, the transcript and the event stream, which cross-check each other. The production grid is the full deployment with its resident nodes and the load they generate, and the bare comparison grid holds only the processes the recovery mechanism needs, so that the measured response belongs to the recovery mechanism itself. The supplementary materials (S.4) list the hardware, the model behind each trial, and the composition of both grids. Two further regimes extend the evaluation over the same three configurations, delegated missions under four fault classes in Section~\ref{sec:missions} and planned changes of one capability inside a continuous session in Section~\ref{sec:changes}, both under mimo-v2.5-pro alone.

\subsection{The Conditions Hold on the Measured Paths}
\label{sec:premises}

The conditions held on every measured path, and the SDK is measured under injected faults against the promise the conditions make. The results show that, under one kill on a single task and under the wider fault model of the delegated missions, every session of the SDK survived, the promise of the conditions held in every faulted task, and both baselines exhibit loss or waste.

Five checks all held. The audit reads each transcript from first entry to last, and the recording order of the store condition holds on every entry. Each transcript replays against the four invariants I1 through I4 with no malformed line. The writer condition was tested by registering the same node id from 100 processes at the same moment, exactly one registration succeeded, the other 99 received an explicit denial, and the table held the single success. E1 was tested under load, 3500 calls in three waves of 50, 100, and 200 concurrent callers with 10 calls per caller, and at the end every call was paired with its answer, with no loss, no duplication, and no misattribution. E3 was tested across 30 rounds of providers arriving and departing among 50, and two observers recorded identical sequences, the event stream alone reconstructing the final table.

\subsection{One Fault on 200 Tasks}
\label{sec:taskcmp}

The SDK kept every session with zero wasted calls and zero lost records, the MCP configuration kept its sessions at the price of 1,212 wasted calls, and the reference lost all 197 faulted sessions. Figure~\ref{fig:outcomes} shows, under DeepSeek V4 Flash, the four quantities in which the cost of a fault appears. Under no fault, task success is comparable across the three configurations, 82.5, 80.0, and 83.0 percent for Logos, the MCP configuration, and the reference under DeepSeek V4 Flash, and 76.0, 77.0, and 72.0 percent under mimo-v2.5-pro.

\begin{figure}
\centering
\small\begin{tabular}{@{}lrrr@{}}\toprule Metric & Logos & MCP & Reference \\\midrule Sessions retained & 197/197 & 197/197 & 0/197 \\ Calls to unavailable endpoint & 0 & 1,212 & 189 \\ Records lost & 0 & 0 & 1,293 \\ Calls redone after recovery & 192 & 125 & 314 \\\bottomrule\end{tabular}
\caption{DeepSeek V4 Flash results for 200 assigned tasks per configuration. One fault opportunity was scheduled per task and was realized in 197 runs per configuration. Session-survival quantities should be read on the realized-fault subset; the remaining quantities report calls and records associated with those runs.}
\label{fig:outcomes}
\end{figure}

The MCP host survives the kill with its session state intact, but the protocol gives no notice of the tool's death, so the model keeps calling the dead endpoint until the watchdog closes the outage window, and the 125 redone calls rebuild the world the server restart erased~\cite{hou2026mcp}.

The single-process reference cannot remount its tool layer, all 197 faulted sessions end at the fault, and under say-once the restarted session delivers only the turns that remain, 44.3 percent of them.

In Logos the fault ends at the tool node, and the 192 redone calls repeat pre-kill calls whose effects the blank restart erased, in 49 of the 197 faulted sessions. In 179 of the 197, 91 percent, the model took the error path and never called the dead endpoint again, and in the remaining 18 the remount completed before the next call.

The pass rates corroborate the mechanism-level quantities under both models. Under DeepSeek V4 Flash, task success runs 60.0, 53.0, and 21.0 percent for Logos, the MCP configuration, and the reference, and the official validator scores 42.5, 36.0, and 1.5 percent. Under mimo-v2.5-pro, task success runs 62.5, 62.5, and 35.0 percent, and the official validator scores 37.5, 44.0, and 1.0 percent. The reference's collapse repeats under both models, a loss of 62.0 points from its no-fault rate under DeepSeek V4 Flash and 37.0 under mimo-v2.5-pro.

Table~\ref{tab:account} lists the costs under DeepSeek V4 Flash and Table~\ref{tab:mimo} under mimo-v2.5-pro, with a wasted call one issued to the dead endpoint during the outage, a lost record one lost with the conversation, and a redone call a post-recovery repeat of a pre-kill call. The weighted effective call rate multiplies the share of calls that are not wasted by the delivered-turn rate, effective time per success includes the outage and the tool timeouts within it, and output tokens per success count the endpoint's output usage over the tasks solved.

\begin{table}[t]
\caption{Costs under the say-once fault regime, DeepSeek V4 Flash as the model, 1 fault per task, 200 tasks each, 2,262, 3,033, and 1,798 calls in total, row definitions in the text.}
\label{tab:account}
\begin{tabular}{@{}p{0.40\linewidth}p{0.155\linewidth}p{0.155\linewidth}p{0.155\linewidth}@{}}
\toprule
Metric & Logos & MCP & DSH \\
\midrule
Wasted calls during outage & 0 & 1,212 & 189 \\
Pre-kill records lost & 0 & 0 & 1,293 \\
Redone calls after recovery & 192 & 125 & 314 \\
Weighted effective call rate & 100\% & 60.0\% & 39.6\% \\
Effective time per success, minutes & 3.14 & 1.40 & 1.54 \\
Output tokens per success & 5{,}040 & 6{,}017 & 19{,}860 \\
\bottomrule
\end{tabular}
\end{table}

\begin{table}[t]
\caption{The same costs under mimo-v2.5-pro, 1 fault per task, 200 tasks each, 1,503, 2,900, and 861 calls in total, the delivered-turn rates 99.6, 100, and 82.4 percent, the row definitions identical to Table~\ref{tab:account}.}
\label{tab:mimo}
\begin{tabular}{@{}p{0.40\linewidth}p{0.155\linewidth}p{0.155\linewidth}p{0.155\linewidth}@{}}
\toprule
Metric & Logos & MCP & DSH \\
\midrule
Wasted calls during outage & 0 & 1,492 & 154 \\
Pre-kill records lost & 0 & 0 & 896 \\
Redone calls after recovery & 64 & 70 & 150 \\
Weighted effective call rate & 99.6\% & 48.6\% & 67.7\% \\
Effective time per success, minutes & 5.40 & 3.53 & 5.14 \\
Output tokens per success & 2{,}102 & 2{,}777 & 3{,}589 \\
\bottomrule
\end{tabular}
\end{table}

Across the two models the identities hold in every faulted task, while the magnitudes change, since how a model responds to the error stream is model behaviour, and the asymmetry among the three configurations is therefore architectural.

\subsection{Connection-Aware Holding as a Host Policy}
\label{sec:hold}

The host policy lifts both cross-process configurations while the identities hold unchanged. A native MCP implementation cannot present a capability-level fault notice of the kind Logos provides~\cite{hou2026mcp}, because the server that would author it is dead, so the policy intercepts connection-class failures, withholds an in-flight call from the model until the tool capability returns, and then reissues it, while the session and the transcript continue unchanged. Applied alike to both configurations, it acts as a control.

On the same 200 tasks the policy lifts the MCP configuration from 36.0 to 69.5 percent on the official validator and from 53.0 to 75.0 percent in task success, and it lifts Logos from 42.5 to 66.0 and from 60.0 to 74.5 percent. With both configurations under the policy the disclosed pass rates are 66.0 against 69.5 on the official validator and 74.5 against 75.0 in task success. They do not decide the comparison, because the pass rates are the one quantity where the two systems come out close. Under the policy Logos still wastes no call and loses no record, and the MCP configuration still repeats 212 calls after recovery, against 186 for Logos.

The mechanism the MCP configuration gained is a client-side state machine built outside its protocol, while in Logos the policy is a corollary of the capability broadcast the bus already carries, with no client-side polling, and while a call is held the model observes nothing of the tool.

The single-process reference admits no such policy, its tool server a child of the session process, with no mounting mechanism and no online broadcast to wait for, and installing either means first building the framework of this paper, after which the system is no longer the reference. Of its 197 faulted tasks, 178 kept the runner process alive past the 60-second interval, and 181 issued calls to the dead tool, 189 calls in all.

\subsection{Why the Session Survives}
\label{sec:guarantee}
\label{sec:response}

\begin{figure}
\centering
\includegraphics[width=\linewidth]{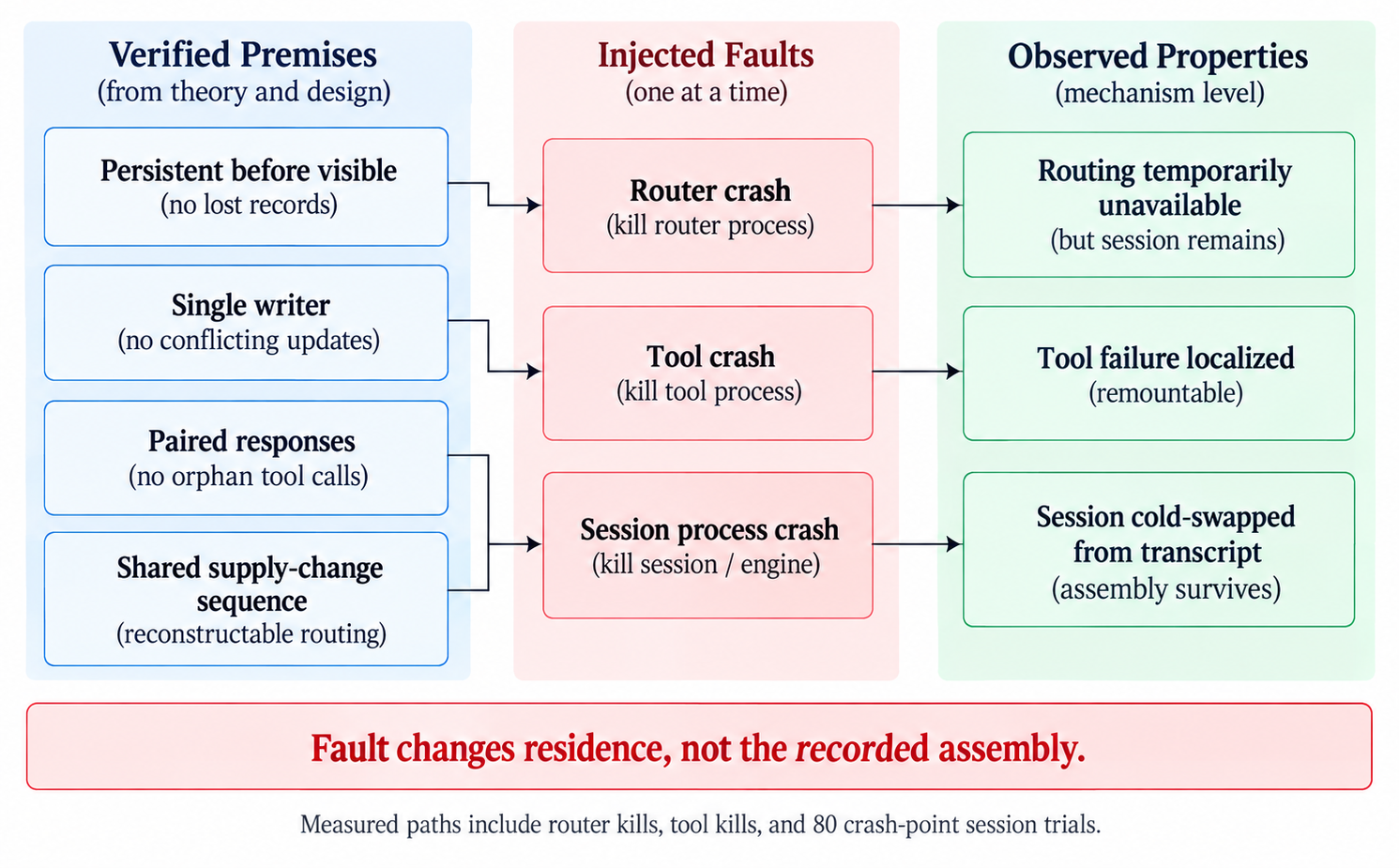}
\caption{Verified disciplines, injected faults, and the properties observed at the mechanism level. The trials cover router kills, tool kills, and the crash-point session trials.}
\label{fig:guarantee}
\end{figure}

Table~\ref{tab:registryprops} maps the four registry properties of Section~\ref{sec:routing} to the trial that attacks each and the result observed, and Figure~\ref{fig:guarantee} traces each discipline through its injected fault to the property observed.

\begin{table}[!t]
\caption{The 4 registry properties of Section~\ref{sec:routing}, each against its trial and its result.}
\label{tab:registryprops}
\begin{tabular}{@{}p{0.15\linewidth}p{0.33\linewidth}p{0.42\linewidth}@{}}
\toprule
Property & Trial & Result \\
\midrule
Externality & the router killed in 20 trials & every node alive, every session running, restoration a median of 858 ms, every trial between 857 ms and 860 ms \\
Name binding & 15 rounds of provider churn, a tool killed in 10 trials & the table equal to the living set in 15 of 15, the provider remounting in 100.5 ms on the production grid \\
Agreement & the observer audit of Section~\ref{sec:premises} & identical sequences across 30 rounds, the event stream alone reconstructing the final table \\
Degradation & a killed tool, a tool coming online during a task & errors to callers in 0.321 ms naming the capability, the first call arriving 8.4 s after the online broadcast with no prompt naming the tool \\
\bottomrule
\end{tabular}
\end{table}

In the single-process form the router and the sessions share one process, so its death ends every session, while in Logos the router carries routing alone. The trial kills the router 20 times, every node stays alive and every session keeps running, and routing resumes after a median of 858 ms, the relaunched router restoring its table by folding the broadcast record. A second trial kills a tool process instead, 10 times, the failure detected in 2.8 ms through the broadcast that follows it.

The same kill costs a remount in Logos and a whole restart in the single-process form. In the single-process configuration an external watchdog watches the session process, detects the kill in 149.5 ms, and restarts the whole stack in 397.9 ms, a median of 547.1 ms per trial for detection and restart together. In Logos the kill lands on a tool node, which remounts in 100.5 ms on the production grid and, over 30 trials, in 41.3 ms on the bare comparison grid, and no other node is affected.

A process can die at any point of a tool call, so the kills are placed at four points of the tool-call cycle, during tool execution, after the return but before persistence, after persistence but before announcement, and after announcement. Eighty sessions, 20 per kill point, run under these kills, and all 80 recover and run to completion. Recovery redoes exactly the calls the transcript has not yet recorded, so the later in the cycle the kill lands, the less there is to redo, four tool rounds at the first two points, three at the third, and two at the fourth, in agreement with Eq.~\eqref{eq:coldswapexec}.

\subsection{The Advantage Holds on Long-Horizon Delegated Missions}
\label{sec:missions}

The mission regime tests both extensions of the conditions at once, engines calling other engines instead of tools, a session spanning a mission of many tasks, and the fault model widening beyond the kill to four classes.

A mission issues one briefing to a main engine, which delegates the work to three scout engines and one analyst engine and delivers four numbers after six to eight steps of delegation. The missions are generated from one seed, so the ground truth is deterministic and success is an exact comparison. Eight missions, five repetitions, and three configurations give 120 records, all under mimo-v2.5-pro.

The seed also decides the faults. Each component carries a 15 percent probability of one fault and takes one of four classes, a kill that stops the component process, a stall that delays the answer by 30 s, a corruption that returns corrupted content, and a silence that returns nothing for 150 s. Across the 40 mission instances the seed injects 17 faults on 14 missions, three kills, six stalls, three corruptions, and five silences, identically placed in all three configurations. The watchdog of 60 s and the call timeout of 120 s take the same values as in the main experiment. Failed-class calls return an error, time out, or return corrupted content, calls on a dead endpoint are counted inside the outage window, and re-dispatches are the main engine tasking the same component again, every assertion checked against the per-component records and the delivered files.

\begin{table}[!t]
\caption{Measurements of the mission regime, 8 missions, 5 repetitions, 3 configurations, triggers identical across configurations, 17 faults on 14 missions, single model, the kill subset carrying 3 paired samples under each configuration.}
\label{tab:mission}
\begin{tabular}{@{}p{0.34\linewidth}p{0.20\linewidth}p{0.20\linewidth}p{0.20\linewidth}@{}}
\toprule
Quantity & Logos & MCP & DSH \\
\midrule
Missions succeeded, of 40 & 40 & 40 & 40 \\
Main sessions alive, of 40 & 40 & 40 & 37 \\
Failed-class calls, total & 11 & 38 & 11 \\
Calls on dead endpoint in outage, total & 0 & 28 & 0 \\
Component re-dispatches, total & 8 & 4 & 13 \\
Wall-clock median, clean, s & 44.0 & 62.4 & 65.1 \\
Wall-clock median, faulted, s & 129.5 & 113.3 & 201.3 \\
Kill subset, main sessions alive, of 3 & 3 & 3 & 0 \\
Kill subset, component return & hot remount 61.0--61.4 s incl. the window & restart 1.2--1.3 s after the window & none, full rerun \\
Kill subset, mission wall-clock, s & 177/322/461 & 113/114/232 & 209/220/285 \\
\bottomrule
\end{tabular}
\end{table}

All missions succeeded in every configuration, and the kill alone separated them. Table~\ref{tab:mission} lists the measurements, and every triple of numbers below keeps the order Logos, the MCP configuration, the reference. The main session survives in all 40 missions in Logos and in the MCP configuration, and in 37 of 40 in the reference.

The three classes that leave the process running behave alike everywhere, because the fault reaches the caller as an ordinary error and the model's retry corrects it in every configuration. The six stalls stretched the medians to 75, 92, and 99 s, the three corruptions were retried by the model itself with 3 calls each, and the five silences answered after the 120 s timeout and succeeded on retry, with medians of 193, 192, and 231 s.

The kills decide the comparison, and the seed provides three paired samples. The main session survived all three kills in Logos and in the MCP configuration, and none of the three in the reference. The reference shares one process between component and runner, so each kill spreads to the session and costs a full rerun of the mission, at 209, 220, and 285 s. The MCP configuration keeps the session in the host, but its protocol sends no notice of the endpoint's death, so during the outage window the model issued 28 calls to the dead endpoint, three to 14 per mission, until the watchdog closed the window. Logos ends the fault at the component process. The routing table answers each call in the window with an explicit refusal naming the capability, and the killed step returns by re-dispatch after a hot remount, completed 61.0 to 61.4 s after the kill, the watchdog window included.

The three configurations therefore absorb the fault at different points, at the call in the MCP configuration, where retries go to the dead endpoint, at the task in Logos, where re-dispatch rebuilds the killed step, and at the mission in the reference, through full reruns. The kill alone removes the process, and the kill alone distinguishes the three configurations, which is what the conditions state, the failure domain at the process boundary. The missions only read state, so the loss of writes at a kill is covered by the single-task experiments of Sections~\ref{sec:taskcmp} and~\ref{sec:guarantee}.

\subsection{The Capability Set Changes while the System Runs}
\label{sec:changes}

The fault of Section~\ref{sec:framework} is itself an unplanned removal, the death of a tool process. This subsection measures the planned form, a deliberate change of supply at runtime, mounting, unmounting, and replacement, under the same three configurations, and then a running system that writes and mounts its own tools.

The unit of measurement is the chain, one continuous session over eight official tasks of the leaderboard, 24 to 25 user turns and 60 to 120 k tokens by its end. At the first tool call of the fourth task the regime applies one planned change to one capability. Mounting brings online a capability the chain has never carried and measures adoption. Unmounting withdraws a capability the session has already used and measures reversion. Replacement exchanges the provider process under the same name and the same schema, a Python process replaced by a Node.js process, and measures continuity of supply. The changed capability is a stateless pure function in every operation, and the outputs of the replacement provider match the original byte for byte (the supplementary materials, S.4).

Each configuration works through its own native path. Logos starts and stops node processes with the router unchanged, the online and offline broadcasts carrying the change to every node. The MCP configuration switches the tool list of an attached server, the server notifies the host, and the host re-fetches the list. The reference registers the capability inside the process and disposes of it through its disposal function. Each pair of operation and configuration runs five change chains and five zero-change control chains, 90 chains in all, under mimo-v2.5-pro alone.

The measured quantities were fixed before the measurements. A probe is a call the evaluation issues to the changed capability every 1.5 s, outside the model's turn, and the co-resident core capability is probed at 3 s through every change window. A change counts as completed when its native path returns success and the host process has not restarted. A chain counts as unmodified when its session loop and system prompt are identical to the controls and its user turns are exactly those of the task scripts. An unmounting counts as reverted when the registration is absent, the provider process has exited, and every later probe is answered with an explicit unavailability naming the capability. At the model layer the regime counts the calls to the changed capability before and after the change, measures the latency from the first user turn that needs a mounted capability to its first call, and compares the official score of every task of a change chain with its paired control.

\subsubsection{The Mechanism Layer}

Every planned change completed without restarting the host, every chain ran through unmodified, and the unmountings reverted in all three configurations, with the mechanism cost differing by configuration. All 45 changes completed through the native path of their configuration, 15 of 15 under each, and all 90 chains ran to their end as unmodified, 30 of 30 under each. All 15 unmountings reverted. The replacement window held no failed probe in Logos and in the reference and one in each MCP chain, and the co-resident core capability answered every probe in every change window. Table~\ref{tab:changes} lists the measurements, including the mechanism cost of each change and the effective latency from the instruction to the first successful probe of the new or replaced capability.

\begin{table}[!t]
\caption{Measurements of the planned-change regime, 5 chains per cell, probe cadence 1.5 s, single model, medians over 5 chains where marked.}
\label{tab:changes}
\begin{tabular}{@{}p{0.40\linewidth}p{0.18\linewidth}p{0.18\linewidth}p{0.18\linewidth}@{}}
\toprule
Quantity & Logos & MCP & DSH \\
\midrule
Changes completed, host never restarting, of 15 & 15 & 15 & 15 \\
Chains unmodified, turns as scripted, of 30 & 30 & 30 & 30 \\
Unmountings reverted, of 5 & 5 & 5 & 5 \\
Failed probes in replacement window, per chain & 0 & 1 & 0 \\
Core-probe failures in change windows & 0 & 0 & 0 \\
First explicit unavailability after unmounting, s, median & 3.35 & 0.99 & 0.57 \\
Mechanism cost, mounting, ms, median & 187 & 410 & 1 \\
Mechanism cost, unmounting, ms, median & 1 & 411 & 1 \\
Mechanism cost, replacement, ms, median & 60 & 2{,}132 & 1 \\
Effective latency, mounting, s, median & 1.19 & 0.80 & 1.23 \\
Effective latency, replacement, s, median & 0.61 & 2.66 & 1.13 \\
Steps to change, and what a step reaches & 1 / 1 / 2, system-wide & 3 / 3 / 3, host-wide & 1 / 1 / 2, process-wide \\
Calls to removed capability, per chain & 0 & 1 & 1--2 \\
Calls to mounted capability, per chain, ground truth 2 & 2 & 2 & 2--3 \\
\bottomrule
\end{tabular}
\end{table}

These results are the registry properties of Section~\ref{sec:routing} at work, the zero failed probes of the replacement window realizing name binding, the explicit unavailability after unmounting realizing degradation, and the mechanism costs following what each configuration must physically do, an in-process registration, a node start, a tool-list switch with a re-fetch, and a process exchange.

The step counts do not grow with the size of the system, and the scale difference appears at removal. Mounting in Logos is one step, a process that starts and registers itself, after which every consumer in the system can call the capability on its next request. Unmounting is one step and replacement two. The MCP configuration takes three steps every time, host-wide. Withdrawing a capability in the reference also reruns the setup of every dependent, a cost that grows linearly with their number (Section~\ref{sec:resultscosts}), while Logos kills one process at a cost independent of that number.

\subsubsection{The Model Layer}

The model adopted every mounted capability, after an unmounting only Logos stopped calling the removed one, and the changes left the tasks themselves intact. Before the unmounting the model had called the capability once in every one of the 15 unmounting chains, all configurations alike. After it, the model called the removed capability zero times in the five Logos chains, once in every MCP chain, and one to two times in the reference. The difference follows from when each configuration changes its tool list. Logos retracts the name with the offline broadcast, and no later request offers it, while the MCP host holds its tool list until the re-fetch and the reference holds its assembly until the next turn, and any attempt to call the removed name lands inside those windows.

After mounting the model called the new capability two times per chain in Logos and in the MCP configuration, and two to three times in the reference, against a ground truth of 2 calls. The first calls arrived 24 to 42, 25 to 39, and 32 to 106 s after the first user turn that needs the capability, a delay set by where the task falls in the chain. The official scores of the change chains and their controls, compared over 32 to 40 paired tasks per combination of operation and configuration, disagree in both directions, with at most 11 tasks discordant on either side.

\subsubsection{A System That Extends Itself}

The capability set can also grow from inside the running system. The system writes a tool, with GPT-5.4 as the model behind the runtime construction of tools, and mounts it on the bus, available to its writer and to the other nodes. In the degradation trial of Table~\ref{tab:registryprops}, the first call to a tool that comes online during a live task arrives 8.4 s after its online broadcast, with no prompt naming the tool, and an extended run repeats the adoption in 10 of 10 trials. Removal, the shrinking side, succeeded with a tracked inverse in 10 of 10 trials. The governance of an agent's own capability set is the question that follows~\cite{gao2026selfevo,shao2026misevolve}.

\subsection{What the Guarantee Costs}
\label{sec:resultscosts}
\label{sec:mechanism}

The guarantee held under every injected fault, and this subsection measures what it costs.

The bus hop stayed negligible against the model call. One bus hop costs a median of 0.215 ms over 10,000 calls, with a 99th percentile of 0.377 ms, a 99.9th percentile of 0.623 ms, and a maximum of 3.045 ms, against an in-process baseline of 0.005 ms, so one hop costs 43 times a local call. The model side costs a median first token of 177 ms and a median full inference of 1,896.8 ms, so one hop is 1/823 of the first token and 1/8822 of the full inference, the separation that the admissibility condition of Section~\ref{sec:mapping} requires. The figure is taken on the loopback path of one host, and a hop across machines adds the round trip of the link, which on a local network still stays far below the first token.

Under no fault the single-process reference and Logos differ by 0.8 percent in makespan, the total time from start to finish, 24,516 ms for the reference against 24,722 ms for Logos. The 0.8 percent gap is the entire saving from centralization, and one fault removes the advantage entirely.

\begin{figure}
\centering
\includegraphics[width=\linewidth]{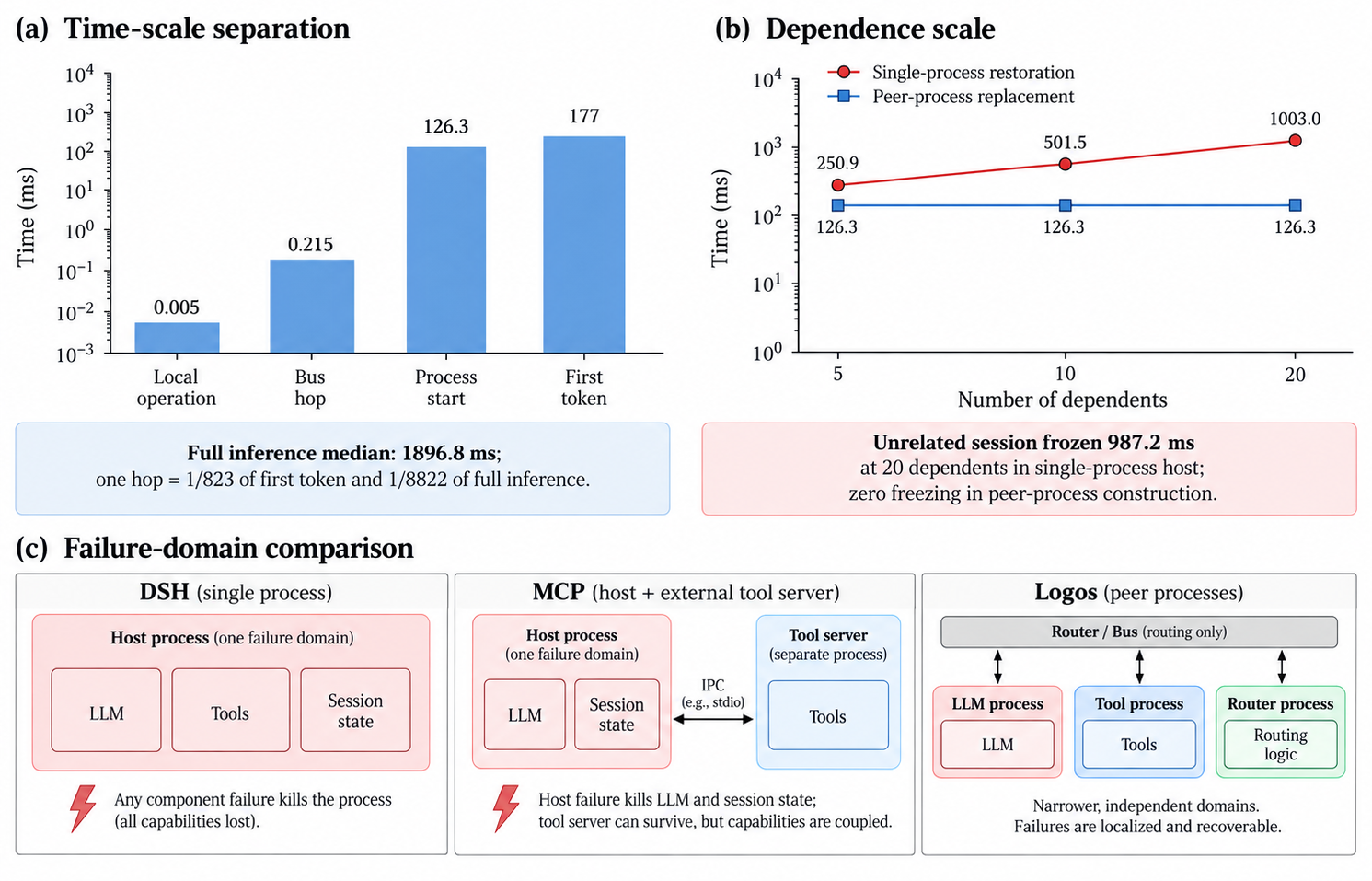}
\caption{The costs of the narrower failure domain. (a) Time-scale separation. (b) The dependence scale of restoration against the constant process start. (c) The failure-domain comparison of the three configurations.}
\label{fig:costs}
\end{figure}

At a change of supply the advantage reverses, restoration after provider withdrawal being serial in the number of dependents in the single-process configuration, each dependent's setup rerunning on the shared event loop. A microbenchmark loads each dependent with 50 ms of synthetic work. The restoration takes 250.9 ms at five dependents, 501.5 ms at 10, and 1003.0 ms at 20, linear in the number of dependents at 50.2 ms per dependent across the three points. An unrelated session in the same host cannot run during the whole restoration and is frozen for 987.2 ms at 20 dependents. Logos incurs one process start of 126.3 ms at the 20-dependent point, and an unrelated session in its own process runs throughout with zero freezing, its scheduling gap a median of 0.2 ms and a maximum of 1.0 ms. A component that loses its remote resource raises an unhandled error. In the single-process host that error propagates past the plugin boundary and terminates the host, while the failing node in Logos returns an explicit error and keeps running. Figure~\ref{fig:costs} shows the time-scale separation, the dependence scale of restoration, and the failure-domain comparison.

The fault itself has a price, and the effective time per success of Tables~\ref{tab:account} and~\ref{tab:mimo} is that price, recovery waiting through the watchdog window and redoing the destroyed work, while the MCP host hides the outage and absorbs it in wasted calls.

\subsection{What the Parts Contribute}
\label{sec:ablation}

The supplementary materials (S.3) read Table~\ref{tab:correspondence} from the removal side, part by part. Seven of the eleven parts keep the assembly host-free, and removing any one of them makes the assembly need a host again. Removing the process boundary of a capability returns every component to one failure domain, and a router that carries business state, a registration that accepts duplicates, or a delivery that reorders its broadcasts each breaks one of the disciplines the conditions rest on. Each removal trial names the property it loses, and the assembly reacquires a host in every one of the seven.

The two baselines are themselves partial removals of this kind, measured under the same fault. The single-process reference keeps an append-only record but holds every capability in one process, which removes the part that places each capability in a process of its own, and it lost every faulted session, the loss that removal predicts. The MCP configuration keeps each tool in a process of its own and moves its servers out, while the assembly, the session record, and the delivery disciplines stay in the host, and it kept its sessions at the price of 1,212 wasted calls on a dead endpoint, the price of the parts it lacks. Each baseline therefore serves as an ablation construction, carrying the advantages of the parts it keeps and paying for the parts it lacks.

The guarantee therefore belongs to the conditions, and the SDK is one fulfillment of them.

\section{Conclusion}
\label{sec:conclusion}

\subsection{Conclusions}

This paper states the conditions under which a plugin-based agent needs no host, and builds the cross-process composition framework that meets them, realized as one SDK named Logos, with the spatiotemporal-composability calculus as the formal support behind the conditions. Four sufficient conditions make host-free composition preserve the session, its record, and its capability set. The SDK meets them on a peer-process bus, a plugin a process, the router a rebuildable routing table, and the session state needed for recovery an append-only transcript owned by no process, from which a replacement process rebuilds a dead session and continues it.

Each claim rests on a measurement. Under one fault on 200 tasks the reference lost its faulted sessions, the MCP configuration kept its sessions but wasted calls on a dead endpoint, and Logos kept every session, with zero wasted calls, zero lost records, and task performance preserved (Section~\ref{sec:resultsguarantee}). Routing returned 858 ms after the router died while the sessions kept running, and the 80 sessions terminated at four points of the tool-call cycle resumed with no repeated effect. All 45 planned changes completed through the native paths, and after an unmounting the removed capability was called zero times in Logos, against once per chain in the MCP configuration (Section~\ref{sec:changes}). One bus hop costs nearly three orders of magnitude less than the first token, and the price of a fault is the watchdog window and the redone work (Section~\ref{sec:resultscosts}).

No process death ends the assembly in these measurements. The bus imposes no language, several LLM execution engines sharing one tool already run in these measurements, and a running system can write a tool and mount it while it serves a session (Section~\ref{sec:changes}).

\begin{figure}
\centering
\includegraphics[width=\linewidth]{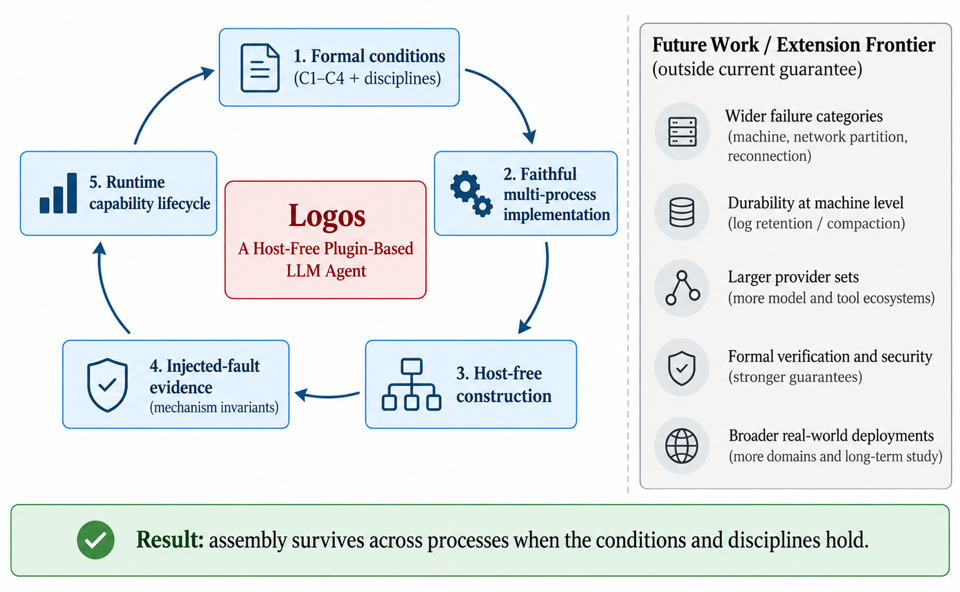}
\caption{The closed loop of the result, from the formal conditions through the faithful implementation, the host-free framework, the injected-fault evidence, and the runtime capability lifecycle. The extension frontier lies outside the current boundary.}
\label{fig:loop}
\end{figure}

A further observation concerns the origin of the parts of this framework. The required functions of Logos overlap to a high degree with the facilities the Robot Operating System already provides~\cite{quigley2009ros,macenski2022ros2}, peer nodes, name-based routing, broadcast delivery, and replaceable processes, and Logos reuses the peer-process bus without building an architecture of its own, which leads to a conjecture, that it may be better to build infrastructure for AI agents from requirements downward, reusing the established facilities of the Internet and the computing stack, than to build it up from frameworks. This paper has answered the question it posed. The conditions are stated with their formal support, the framework is built, one SDK is measured, and the failure domain narrows to the process that carries the fault. Figure~\ref{fig:loop} closes the loop of the result and marks the extension frontier beyond the current boundary.

\subsection{Limitations and Future Work}
\label{sec:scope}

\textit{The fault form and the scope of the measurements.} One fault is injected as a kill of a process, a form that covers every fault which ends in process termination, crash, unhandled exception, memory exhaustion, watchdog termination, and an external kill, the loss of everything inside the process~\cite{schlichting1983failstop}. Recovery consults only the transcript and remains independent of the cause, so the measured recovery properties extend to every fault of the class. Two further classes lie outside this form, the class in which a process keeps running while its behaviour departs from its specification, whose bus-level forms belong to the formal semantics of loss, partition, and reconnection that the next step develops~\cite{lamport1982byzantine}, and machine loss, because the append of the transcript is not forced to disk.

\textit{The boundary of what is stated.} The stated boundary covers assembly and removal with their records, and the store condition, the writer condition, and the delivery constraints are audited on the measured paths. Effects on the external world stay outside the boundary, untracked and unrecovered, and withholding and compensation are their only dispositions~\cite{dsh2026spatiotemporal}. After a cold swap the rebuilt view equals the view the failed process held, and whether the model answers identically after a recovery lies outside the boundary.

\textit{The fencing of the writer condition.} The single-writer condition is fenced at the connection. A second live connection under one ID is refused at registration, a successor claims the ID only after the connection that held it has died, and a displaced connection has every later registration denied.

\textit{Log growth.} The transcript is the complete record of every step, and the archived event stream of the faulted run holds 509,506 events of 20 kinds. Retention and compaction policies belong to the deployment.

\begin{figure}[!ht]
\centering
\includegraphics[width=\linewidth]{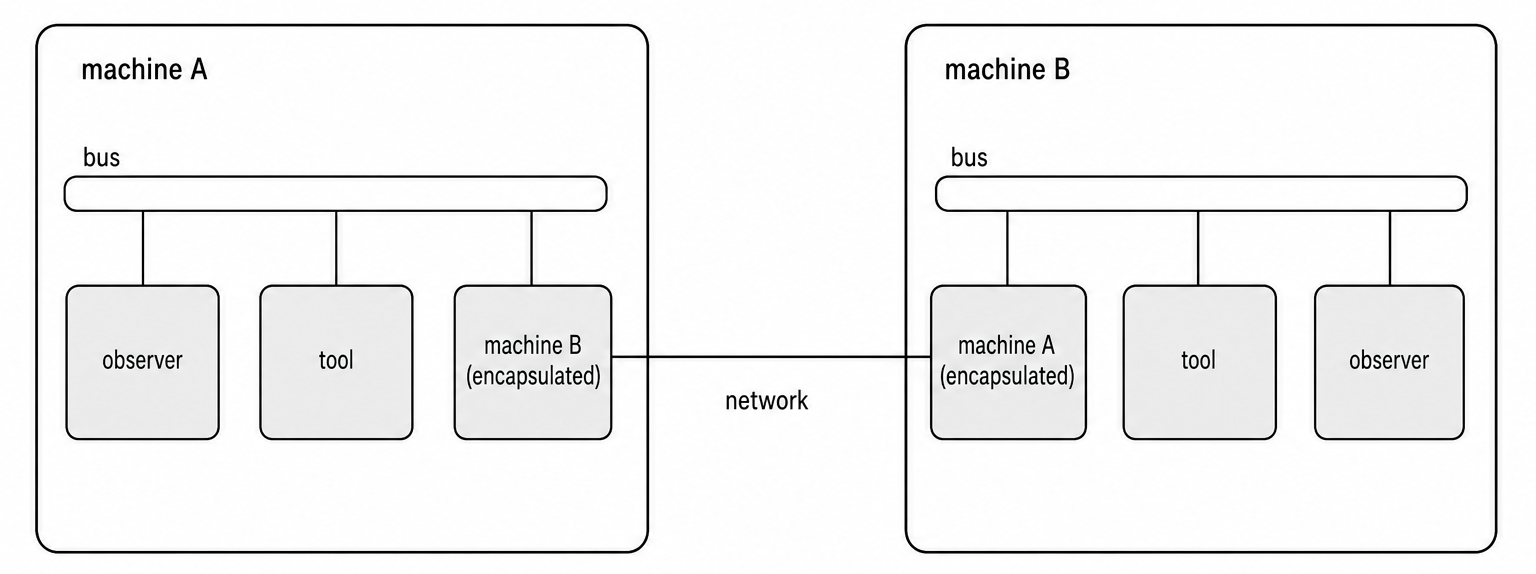}
\caption{Two machines joined by a network, each one node on the other's bus. The far machine is mounted as a node of the same form as every other, and the machine boundary lies below the node.}
\label{fig:crossmachine}
\end{figure}

\textit{Deployment beyond one machine.} The next step is a measured cross-machine deployment, with a formal semantics of loss, partition, and reconnection (Figure~\ref{fig:crossmachine})~\cite{gilbert2002brewer}.

\section*{CRediT authorship contribution statement}

\textbf{Hanzhang Jia:} Conceptualization, Methodology, Software, Writing - original draft.
\textbf{Liheng Zeng:} Software, Validation, Investigation.
\textbf{Hao Cheng:} Validation, Data curation.
\textbf{Yi Gao:} Resources, Visualization.
\textbf{Bo Ma:} Validation, Writing - review \& editing.
\textbf{Guolong Xu:} Supervision, Writing - review \& editing.

\section*{Declaration of competing interest}

The authors declare that they have no known competing financial interests or personal relationships that could have appeared to influence the work reported in this paper.

\section*{Data availability}

The evaluation code and the measurement records are available, and the reference implementation is public~\cite{dsh2026repo}.

\bibliographystyle{elsarticle-num}
\bibliography{refs}

\end{document}